\def\year{2022}\relax
\documentclass[letterpaper]{article} %
\usepackage{aaai22}  %
\usepackage{times}  %
\usepackage{helvet}  %
\usepackage{courier}  %
\usepackage[hyphens]{url}  %
\usepackage{graphicx} %
\usepackage{natbib}  %
\usepackage{caption} %
\DeclareCaptionStyle{ruled}{labelfont=normalfont,labelsep=colon,strut=off} %
\usepackage{algorithm}
\usepackage{algorithmic}

\usepackage{newfloat}
\usepackage{listings}
\floatstyle{ruled}
\newfloat{listing}{tb}{lst}{}
\floatname{listing}{Listing}

\usepackage{pifont}%
\newcommand{\cmark}{\ding{51}}%
\newcommand{\xmark}{\ding{55}}%
\usepackage{soul}

\usepackage{amsmath}

\usepackage{multirow}
\usepackage{booktabs}
\usepackage{subcaption}

\usepackage[hidelinks]{hyperref}
\hypersetup{colorlinks,linkcolor={black},citecolor={black},urlcolor={black}}

\usepackage{lipsum}
\usepackage{fixltx2e}
\usepackage{romannum}

\usepackage{float}
\usepackage{diagbox} 
\usepackage{makecell}

\usepackage[bottom]{footmisc}

\title{Adversarial Purification through Representation Disentanglement}
\author{
    Tao Bai, Jun Zhao, Lanqing Guo, Bihan Wen
}
\affiliations{

    \textsuperscript{\rm 1}Nanyang Technological University\\
    \{bait0002, junzhao, lanqing001, bihan.wen\}@ntu.edu.sg
}

\begin{document}

\maketitle

\begin{abstract}
Deep learning models are vulnerable to adversarial examples and make incomprehensible mistakes, which puts a threat on their real-world deployment.
Combined with the idea of adversarial training, preprocessing-based defenses are popular and convenient to use because of their task independence and good generalizability.
Current defense methods, especially purification, tend to remove ``noise" by learning and recovering the natural images. However, different from random noise, the adversarial patterns are much easier to be overfitted during model training due to their strong correlation to the images.
In this work, we propose a novel adversarial purification scheme by presenting disentanglement of natural images and adversarial perturbations as a preprocessing defense.
With extensive experiments, our defense is shown to be generalizable and make significant protection against unseen strong adversarial attacks.
It reduces the success rates of state-of-the-art \textbf{ensemble} attacks from \textbf{61.7\%} to \textbf{14.9\%} on average, superior to a number of existing methods.
Notably, our defense restores the perturbed images perfectly and does not hurt the clean accuracy of backbone models, which is highly desirable in practice.
\end{abstract}

\section{Introduction}
Adversarial examples have seriously threatened the application of deep learning models, which are mainly crafted by adding malicious perturbations to natural images~\cite{DBLP:journals/corr/SzegedyZSBEGF13,DBLP:journals/corr/GoodfellowSS14,DBLP:conf/iclr/MadryMSTV18,DBLP:conf/cvpr/DongLPS0HL18,Dong_2019_CVPR}.
To enhance and harden the reliability of deep learning models, a large number of researchers are devoted to developing defenses to counter adversarial examples, such as adversarial training~\cite{DBLP:conf/iclr/MadryMSTV18,pmlr-v97-zhang19p} and input preprocessing defenses~\cite{DBLP:journals/corr/abs-1711-00117,xie2018mitigating,DBLP:conf/cvpr/PrakashMGDS18,DBLP:conf/cvpr/LiaoLDPH018,Naseer_2020_CVPR}.
Adversarial training methods are usually task-dependent, computationally expensive, and sacrifice models' performances on clean data~\cite{Naseer_2020_CVPR}.
In contrast, input preprocessing defenses are scalable and task-agnostic.

Now we are discussing the design of preprocessing methods. 
While early input preprocessing methods~\cite{DBLP:journals/corr/abs-1711-00117,xie2018mitigating,DBLP:conf/cvpr/PrakashMGDS18} are less robust to strong iterative attacks~\cite{Dong_2019_CVPR}, the recently proposed purification defenses~\cite{DBLP:conf/cvpr/LiaoLDPH018,Naseer_2020_CVPR} show more promising results by regarding the adversarial perturbation removal as a special case of denoising tasks.
Popular deep denoising schemes~\cite{zhang2017beyond,liu2018image} focus on learning the underlying image structures from the clean corpus. 
Thus, as the solver of the ill-posed inverse problem, i.e., denoising, the learned deep image prior can be applied to remove ``noise" that is assumed random and uncorrelated to the clean images.
In contrast, such assumption is no longer valid in adversarial purification, as the perturbations are generated based on the natural data, thus NOT random but highly correlated to the natural images.
Most attack methods take the images as inputs and accumulate the gradients of models w.r.t inputs to craft adversarial examples.
Thus, as existing purification defense algorithms all reply on fitting the mapping from adversarial samples to natural images, they inevitably overfit the adversarial patterns when constructing the deep image prior.
We argue that such an overfitting issue by overlooking the fundamental differences between adversarial perturbations and random noise would limit the performances of existing purification-based defenses.

\begin{figure}[t]
\centering
\includegraphics[width=\linewidth]{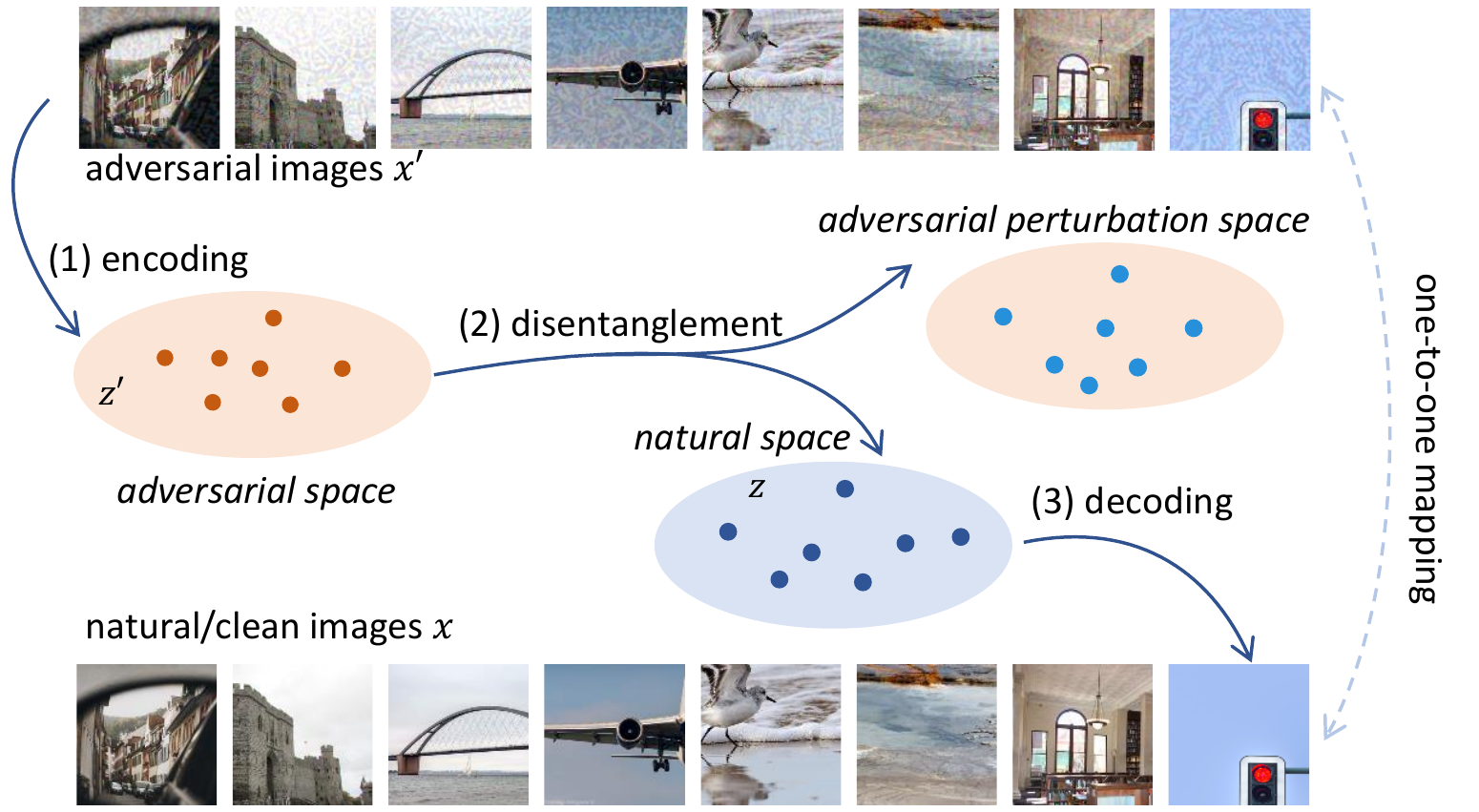}
\caption{
The pipeline of the proposed RDP for adversarial purification.
}
\label{fig:motivation}
\end{figure}

In this work, different from all existing methods, we tackle the adversarial perturbation challenge by learning to disentangle perturbation and natural images, based on which we propose a novel adversarial example purification method, called \textbf{R}epresentation \textbf{D}isentangled \textbf{P}urification (\textbf{RDP}).
Concretely, RDP disentangles the latent vectors of natural images from adversarial latent space and recovers natural images in an end-to-end manner.
Figure~\ref{fig:motivation} shows the pipeline of the proposed RDP.
RDP shows the best performances against various strong adversarial attacks through extensive experiments, compared to the state-of-the-art~(SOTA) preprocessing-based defenses.
Our contributions are summarized as follows:
\begin{itemize}
  \item We propose a novel adversarial defense scheme named RDP, which effectively purifies the adversarial examples by disentangling natural image representations from adversarial perturbations.
  \item The RDP model is trained in a self-supervised manner with no assumption of potential adversarial attacks. Notably, RDP can be turned into a dynamic defense against strong white-box attacks.
  \item Extensive experimental results against various strong attacks on ImageNet demonstrate the efficacy and superiority of RDP.
\end{itemize}

\section{Related Works}
\subsection{Adversarial Attacks}
Since \citeauthor{DBLP:journals/corr/SzegedyZSBEGF13}~\shortcite{DBLP:journals/corr/SzegedyZSBEGF13} for the first time revealed the vulnerability of deep learning models on adversarial examples, numerous attack methods are proposed to generate imperceptible adversarial perturbations.
\citeauthor{DBLP:journals/corr/GoodfellowSS14}~\shortcite{DBLP:journals/corr/GoodfellowSS14} proposed an one-step attack: Fast Gradient Sign Method (FGSM) to generate adversarial examples quickly;
Later, the multi-step FGSM (I-FGSM) is developed in \cite{DBLP:journals/corr/KurakinGB16} to generate strong attacks.
Some similar attacks are C\&W attack~\cite{carlini2017towards}, JSMA~\cite{papernot2016limitations}, and PGD attack~\cite{DBLP:conf/iclr/MadryMSTV18}.
To enhance the transferability and practicality of strong attacks, Momentum is introduced in the process of generation, called MI-FGSM~\cite{DBLP:conf/cvpr/DongLPS0HL18}.
More recently, \citeauthor{Xie_2019_CVPR}~\shortcite{Xie_2019_CVPR} proposed to augment the input with randomized scaling and padding to boost the transferability, the attack of which is called DI-FGSM; 
\citeauthor{Dong_2019_CVPR}~\shortcite{Dong_2019_CVPR} and \citeauthor{Lin2020Nesterov}~\shortcite{Lin2020Nesterov} utilized the translation-invariant and scale-invariant properties of convolutional neural networks (CNNs), and developed corresponding TI-FGSM and SI-FGSM attacks respectively.
All these attacks are generated as a function of the input images.

\subsection{Adversarial Defenses}
One widely recognized defense method is called adversarial training~\cite{DBLP:conf/iclr/MadryMSTV18,pmlr-v97-zhang19p}, which involves adversarial examples during training, and many variants of adversarial training have been proposed~\cite{bai2021recent}.
Though effective, there are many challenges of adversarial training remaining to solve, like high training cost, task dependency, and the trade-off between accuracy and adversarial robustness for adversarial trained models~\cite{Naseer_2020_CVPR, bai2021recent}, which hinder the broad application of adversarially trained models.
Compared to adversarial training, another line of research relies on image restoration techniques to remove adversarial perturbations before feeding images to downstream tasks.
\citeauthor{DBLP:journals/corr/abs-1711-00117}~\shortcite{DBLP:journals/corr/abs-1711-00117} used JPEG compression and Total Variation Minimization (TVM) to remove adversarial perturbations.
\citeauthor{xu2017feature}~\shortcite{xu2017feature} proposed bit reduction (Bit-Red), which is designed to do feature squeezing to remove adversarial effects.
\citeauthor{xie2018mitigating}~\shortcite{xie2018mitigating} preprocessed the input by Random resizing and Padding (R\&P) to mitigate the effects of adversarial examples.
\citeauthor{mustafa2019image}~\shortcite{mustafa2019image} employed super-resolution (SR) to map off-the-manifold adversarial samples back to the manifold of natural images.
\citeauthor{jia2019comdefend}~\shortcite{jia2019comdefend} utilized deep models to compress input images and remove adversarial patterns, which is named as ComDefend.
\citeauthor{liu2019feature}~\shortcite{liu2019feature} proposed Feature distillation (FD), which is a JPEG-based compression framework for defense purposes.
\citeauthor{DBLP:conf/cvpr/LiaoLDPH018}~\shortcite{DBLP:conf/cvpr/LiaoLDPH018} adopted the denoiser with high-level representation guidance, which is named HGD.
R\&P and HGD are the top-2 solutions in NIPS 2017 Adversarial attacks and defenses competition~\cite{kurakin2018adversarial}.
NRP~\cite{Naseer_2020_CVPR}, the derivative work of HGD, employs a self-supervised way to generate adversarial examples for training denoisers.
Energy-Based Models (EBM) trained with Markov-Chain Monte-Carlo (MCMC)~\cite{yoon2021adversarial,Grathwohl2020Your} show the efficacy for purifying adversarial examples from toy datasets.
A large number of MCMC steps during training make these methods costly and impractical when applied to large datasets.
The online purification method proposed by~\cite{shi2021online} requires an accompanied auxiliary task for training backbone models, which is task-dependant thus not considered in this paper.

\subsection{Disentanglement for Adversarial Robustness}
Disentanglement is designed to model the independent representations of data with variations, which has a wide range of applications in image translation~\cite{liu2018unified} and domain adaptation~\cite{peng2019domain}. 
In the field of adversarial defenses, some very recent works have demonstrated the efficacy of disentanglement in robust representation learning~\cite{willetts2020disentangling,yang2021adversarial,gowal2020achieving}.
To the best of our knowledge, no prior works apply disentanglement to adversarial purification.
We are the first to exploit disentangling the natural representations from adversarial examples and restore their natural counterparts.

\section{Methodology}
In this section, we first elaborate on the architecture and the objective functions of RDP, then explain the motivation and advantages of the two-branch design of RDP.

\subsection{RDP Framework}
\begin{figure*}[ht]
\centering
\includegraphics[width=0.95\linewidth]{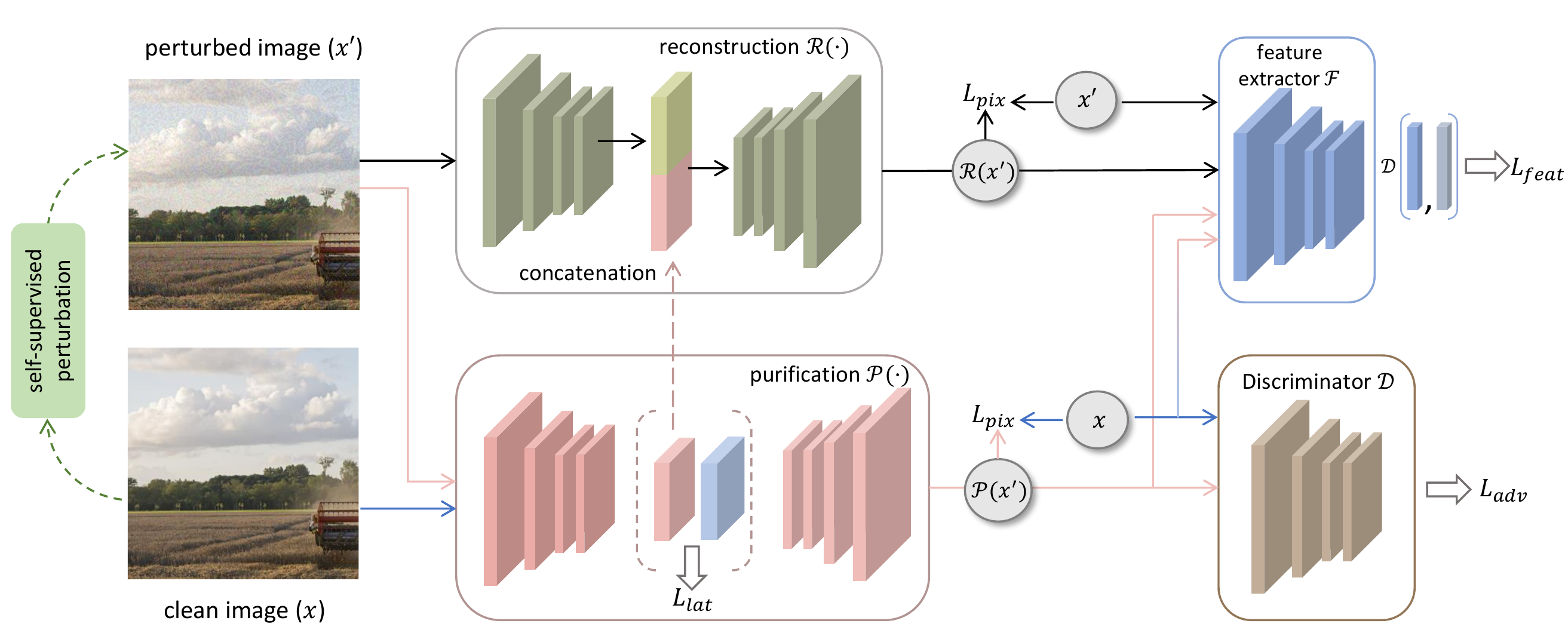}
\caption{Illustration of proposed RDP network architecture.
}
\label{fig:arch}
\end{figure*}

The design of our RDP architecture is depicted in this section.
As shown in Figure~\ref{fig:arch}, RDP consists of a purifier $\mathcal{P}$, a reconstructor $\mathcal{R}$, a discriminator $\mathcal{D}$ and a feature extractor $\mathcal{F}$. 
The upper and lower branches in DRP are called the purification and reconstruction branches, respectively.
The proposed RDP aims to learn a purifier $\mathcal{P}$ (purification branch) to purify adversarial patterns and restore natural images ($x$) from their adversarial examples ($x^{\prime}$), and utilize the reconstructor $\mathcal{R}$ (reconstruction branch) to reconstruct the perturbed images with the natural latent representations as the additional inputs.
In this way, natural latent representations are disentangled from adversarial latent representations and used to recover purified images.
The purified images are then discriminated by the discriminator $\mathcal{D}$ to ensure they are mapped onto the manifold of natural images.
With the pre-trained feature extractor $\mathcal{F}$, we utilize the power of self-supervision for crafting adversarial examples, which makes our defense task-agnostic and is proved to be effective in~\cite{Naseer_2020_CVPR}.
During training, given natural images $x$, adversarial examples $x^{\prime}$ are crafted by maximizing the feature distortions of the feature extractor $\mathcal{F}$, which is expressed as
\begin{equation}
\begin{split}
\max _{\boldsymbol{x}^{\prime}} \boldsymbol{d}\left(\mathcal{F}_{n}(\boldsymbol{x}),\mathcal{F}_{n}\left(\boldsymbol{x}^{\prime}\right)\right) \text { s.t. }\left\|\boldsymbol{x}-\boldsymbol{x}^{\prime}\right\|_{\infty} \leq \epsilon,
\end{split}
\label{eqn:feature distortion}
\end{equation}
where $n$ is layer number of $\mathcal{F}$ used for feature extraction, $\epsilon$ is the maximum of perturbations under $l_{\infty}$ distance, and the function of $\boldsymbol{d}(\cdot)$ is to calculate the distance between two inputs.
We use project gradient descent~\cite{DBLP:conf/iclr/MadryMSTV18} to solve the Eq.~(\ref{eqn:feature distortion}) and obtain $x^{\prime}$.
Then $x$ and $x^{\prime}$ are fed into RDP simultaneously.
The training scheme is summarized in Algorithm~\ref{alg:algorithm}.

\subsection{Training Loss}
For the purification branch, $x^{\prime}$ is expected to be mapped into the natural image manifold.
The purified image $\mathcal{P}(x^{\prime})$ should be aligned to the natural image $x$ to a feasible extent.
Thus, we designed a hybrid loss function for $\mathcal{P}$, which consists of \textbf{latent representation loss}, \textbf{feature loss}, \textbf{pixel loss}, and \textbf{adversarial loss} and is well explained in the following.

RDP is to disentangle the natural representations from adversarial examples and restore the natural counterparts.
The restored images should be similar to the clean images as much as possible.
We design the feature loss $\mathcal{L}_{feat}$ and pixel loss $\mathcal{L}_{pix}$ for restoring images from contents to styles, which are expressed as
\begin{equation}\begin{aligned}
\mathcal{L}_{feat}^{\mathcal{P}} = \left\|\mathcal{F}_{n}(\mathcal{P}(x^{\prime})) - \mathcal{F}_{n}(x)\right\|_{1}
\end{aligned}
\end{equation}
and 
\begin{equation}\begin{aligned}
\mathcal{L}_{pix}^{\mathcal{P}} = \left\|\mathcal{P}(x^{\prime}) - x)\right\|_{1}.
\end{aligned}
\end{equation}
We add latent representation loss $\mathcal{L}_{lat}$ for regularizing the latent representations extracted by the encoder, which is expressed as 
\begin{equation}\begin{aligned}
\mathcal{L}_{lat} = \left\|\mathcal{P}_{enc}(x^{\prime}) - \mathcal{P}_{enc}(x)\right\|_{1},
\end{aligned}
\end{equation}
where $\mathcal{P}_{enc}(\cdot)$ represents the disentangled latent representations.
Lastly, we use the adversarial loss $\mathcal{L}_{adv}$ to push the restored images to the manifold of natural images.
We use the relativistic average GAN objective here for better convergence~\cite{yadav2020relativistic,jolicoeur2018relativistic,Naseer_2020_CVPR}, which is 
\begin{equation}
\mathcal{L}_{adv}=-\log \left(\sigma\left(\mathcal{D}_{\phi}\left(\mathcal{P}\left(\boldsymbol{x}^{\prime}\right)\right)-\mathcal{D}_{\phi}(\boldsymbol{x})\right)\right),
\end{equation}
where $\sigma$ is the sigmoid function.
The overall loss function of purification branch is 
\begin{equation}
\mathcal{L}_{\mathcal{P}}= \mathcal{L}_{feat}^{\mathcal{P}} + \alpha_{1} \cdot \mathcal{L}_{pix}^{\mathcal{P}} + \alpha_{2} \cdot \mathcal{L}_{lat} + \alpha_{3} \cdot \mathcal{L}_{adv}.
\label{eqn:lp}
\end{equation}

On the other hand, the reconstruction branch is to reconstruct adversarial examples.
It is worthy to note that the reconstructor $\mathcal{R}$ takes the latent representations extracted by $\mathcal{P}$ as input as well.
Ideally, the natural latent representations from $\mathcal{P}$ are concatenated with the latent representations extracted by the encoder of $\mathcal{R}$, forming the integral adversarial latent representations for reconstruction.
As such, the reconstruction and the purification branches complete the processing of disentanglement together.
Thus, the reconstructed images $\mathcal{R}(x^{\prime})$ are expected to be similar to $x^{\prime}$ from the \textbf{feature} and \textbf{pixel} level, the losses of which are expressed as
\begin{equation}\begin{aligned}
\mathcal{L}_{feat}^{\mathcal{R}} = \left\|\mathcal{F}_{n}(\mathcal{R}(x^{\prime})) - \mathcal{F}_{n}(x^{\prime})\right\|_{1}
\end{aligned}
\end{equation}
and 
\begin{equation}\begin{aligned}
\mathcal{L}_{pix}^{\mathcal{R}} = \left\|\mathcal{R}(x^{\prime}) - x^{\prime})\right\|_{1}.
\end{aligned}
\end{equation}
Similarly, the overall loss for the reconstruction branch is 
\begin{equation}
\mathcal{L}_{\mathcal{R}}= \mathcal{L}_{feat}^{\mathcal{R}} + \beta \cdot \mathcal{L}_{pix}^{\mathcal{R}}.
\label{eqn:lr}
\end{equation}

\begin{algorithm}[t]
\caption{RDP training}
\label{alg:algorithm}
\textbf{Input}: training data $\mathcal{X}$, Purifier $\mathcal{P}$, Reconstructor $\mathcal{R}$, feature extractor $\mathcal{F}$, discriminator $\mathcal{D}$, perturbation budget $\epsilon$, loss functions $\mathcal{L}_{\mathcal{P}}$ and $\mathcal{L}_{\mathcal{R}}$\\
\textbf{Initialization}: $\mathcal{P}$, $\mathcal{R}$ and $\mathcal{D}$
\begin{algorithmic}[1] %
\WHILE{$\mathcal{P}$ doesn't converge}
\STATE Sample mini-batch of data $x \sim \mathcal{X}$.
\STATE Obtain adversarial counterparts $x^{\prime}$ within the perturbation budget $\epsilon$ by solving Eq.~(\ref{eqn:feature distortion}).
\STATE Feed $x$ and $x^{\prime}$ into $\mathcal{P}$, extract $V_{x^{\prime}}$ (the latent representations of $x^{\prime}$), and calculate $\mathcal{L}_{\mathcal{P}}$
(Eq.~(\ref{eqn:lp})).
\STATE Feed $x^{\prime}$ and $V_{x^{\prime}}$ into $\mathcal{R}$ and calculate $\mathcal{L}_{\mathcal{R}}$ (Eq.~(\ref{eqn:lr})).
\STATE Update $\mathcal{P}$ and $\mathcal{R}$ to minimize $\mathcal{L}_{\mathcal{P}}$ and $\mathcal{L}_{\mathcal{R}}$, respectively.
\STATE Update $\mathcal{D}$ to classify $x$ from $\mathcal{P}(x^{\prime})$.
\ENDWHILE
\STATE \textbf{return} $\mathcal{P}$
\end{algorithmic}
\end{algorithm}

\subsection{Motivation of RDP Design}
\begin{figure}[t]
\centering
\begin{subfigure}{0.24\linewidth}
\includegraphics[width=\linewidth]{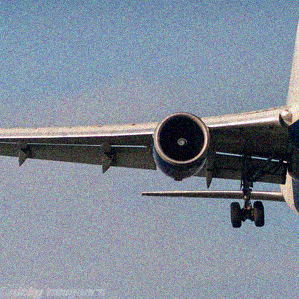}
\caption{}
\label{fig:noise diff-a}
\end{subfigure}
\begin{subfigure}{0.24\linewidth}
\includegraphics[width=\linewidth]{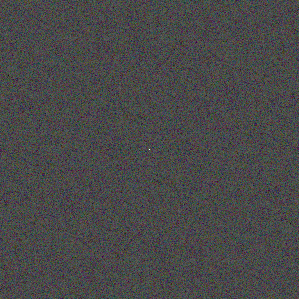}
\caption{}
\label{fig:noise diff-c}
\end{subfigure}
\begin{subfigure}{0.24\linewidth}
\includegraphics[width=\linewidth]{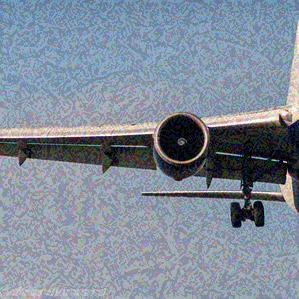}
\caption{}
\label{fig:noise diff-b}
\end{subfigure}
\begin{subfigure}{0.24\linewidth}
\includegraphics[width=\linewidth]{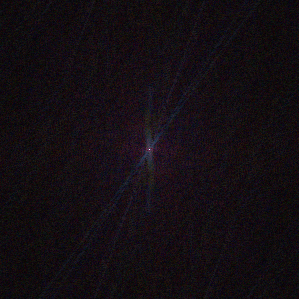}
\caption{}
\label{fig:noise diff-d}
\end{subfigure}
\caption{
Differences between Gaussian noise and adversarial perturbations from the perspective of frequency domain. (a) and (c) are samples with Gaussian noise and adversarial perturbations, respectively. (b) and (d) plots the magnitudes of the noise and perturbation frequency spectrum in (a) and (c), respectively.
}
\label{fig:noise diff}
\end{figure}

The design of RDP is to model both the natural images and their adversarial perturbations, which are different from random noise in two aspects.
First, the distribution of adversarial perturbations is unknown.
Random noise, due to Central Limit Theorem, is usually assumed to be under the Gaussian distribution~\cite{6751276,zhu2016blind}.
In contrast, as stated above, adversarial perturbations are dependant to images, which can be seen as a function of input images.
The distribution of adversarial perturbations is difficult to model explicitly and remains unclear to us~\cite{DBLP:conf/nips/DongDP0020}.
Second, adversarial perturbations are correlated to image contents.
From the perspective of the frequency domain, adversarial perturbations tend to have more high-frequency components that are mixed with the image contents while Gaussian noise does not (see Figure~\ref{fig:noise diff}).
As discovered in~\cite{ulyanov2018deep}, deep neural networks tend to learn the structural contents from corrupted images.
The single-branch denoising methods inevitably would be overfitting to adversarial patterns during training, hurting the restoration of corrupted images.

By design, our two-branch RDP can handle these challenges.
It is known that adversarial examples are crafted by adding adversarial perturbations to clean images.
In our approach, as shown in Figure~\ref{fig:arch}, we use the purification branch to restore clean images from adversarial inputs, and the reconstruction branch to reconstruct the adversarial inputs.
On the one hand, the purification branch learns the distribution of natural images.
On the other hand, the two branches are connected through latent representations.
The adversarial images are recovered from the concatenated latent representations so that the distribution of adversarial perturbations is captured implicitly by RDP,
which makes RDP superior to single-branch methods.

\section{Experiments}
\subsection{Implementation details}\label{sec:exp imp}
\paragraph{Model Architecture} There are mainly four sub-models in RDP.
The architectures of purifier $\mathcal{P}$ and the reconstructor $\mathcal{R}$ are adopted from~\cite{wang2019edvr,kupyn2018deblurgan}, which consist of a convolutional layer and multiple basic blocks.
Like~\citeauthor{Naseer_2020_CVPR}~(\citeyear{Naseer_2020_CVPR}) did, we remove the skip-connection to avoid introducing adversarial patterns to outputs.
For the feature extractor $\mathcal{F}$ and the discriminator $\mathcal{D}$, their architectures are based on the VGG network, which is composed of five convolutional layers and one fully connected layer.
Note that $\mathcal{F}$ is pre-trained on ImageNet~\cite{ILSVRC15} and fixed during training.
\paragraph{Training Details}
We implement our models and experiments with Python 3.8 and Pytorch 1.7~\cite{paszke2019pytorch} on three NVIDIA GTX 2080 Ti GPUs.
For training, we randomly extract 25K images from the training data of ImageNet~\cite{ILSVRC15} (25 images per class) as our training set.
Adversarial examples are generated in a self-supervised way with $\epsilon=16$ and fed into RDP with their corresponding natural images.
During training, we use randomly cropped images whose size is 128 × 128 × 3. 
The batch size is set to 48.
Learning rates $\mathcal{R}$, $\mathcal{P}$ and $\mathcal{D}$ for are 0.0001.
Hyperparameters are set to be $\alpha_{1,2,3} = \{10^{-2}, 10^{-1},  10^{-4}\}$ and $\beta = 10^{-4}$, respectively.
\paragraph{Evaluation Details}
For evaluation, we mainly use NIPS 2017 DEV dataset~\cite{kurakin2018adversarial}, which contains $1,000$ images sampled from ImageNet.
In our experiments, we study five models naturally pretrained on ImageNet: Inceptionv3 (Inc-v3)~\cite{szegedy2016rethinking}, Inceptionv4 (Inc-v4)~\cite{szegedy2017inception}, ResNet v2-101 (ResNet-101)~\cite{he2016identity} , ResNet v2-152 (ResNet-152) and Inception-ResNet-v2 (IncRes-v2)~\cite{szegedy2017inception}, and one adversarial trained model IncRes-v2\textsubscript{ens}~\cite{DBLP:conf/iclr/TramerKPGBM18}.
Using the author-released codes, we evaluate our defense with various attacks like FGSM\footnote{\label{fn:tim}https://github.com/dongyp13/Translation-Invariant-Attacks}, MI-FGSM (iteration=10, momentum = 1)\footref{fn:tim}, DIM (probability = 0.7)\footnote{https://github.com/cihangxie/DI-2-FGSM/}, TI-DIM\footref{fn:tim}, and SINI-TIDIM (scale = 5)\footnote{https://github.com/JHL-HUST/SI-NI-FGSM}.
Note that we only use the purification branch of RDP for evaluation.

\begin{table}[tp]
\centering
\caption{Error rates (\%, lower is better) of different defense methods against various attacks ($\epsilon=16$). Other defenses are implemented based on author-released codes.}
\label{tab:various attacks}
\resizebox{\columnwidth}{!}{%
\begin{tabular}{cccccccc}
\toprule
&  & \multicolumn{5}{c}{Attacks}                                                                 &  \\ \cmidrule(r){3-8} 
          & \rotatebox{90}{Defense} & \rotatebox{90}{FGSM}         & \rotatebox{90}{MI-FGSM}          & \rotatebox{90}{DIM}          &
\rotatebox{90}{TI-DIM}              & \rotatebox{90}{SINI-TIDIM}           &   \rotatebox{90}{Average} \\ \midrule
\multirow{10}{*}{\rotatebox{90}{Inc-v3}}          & JEPG                     & 19.4         & 20.6          & 31.8          & 39.6          & 44.1          & 31.1          \\
                                  & FD                       & 27.5         & 26.8          & 33.4          & 47.8          & 42.8          & 35.7          \\
                                  & Bit-Red                      & 8.1          & 10.5          & 13.4          & 34.2          & 20.5          & 17.3          \\
                                  & TVM                      & 10.0         & 13.8          & 18.6          & 37.8          & 27.7          & 21.6          \\
                                  & SR                       & 28.0         & 28.9          & 44.3          & 40.8          & 43.7          & 37.1          \\
                                  & R\&P                     & 8.9          & 11.8          & 17.9          & 41.2          & 21.6          & 20.3          \\
                                  & ComDefend                & 22.5         & 22.5          & 27.3          & 44.3          & 37.3          & 30.8          \\
                                  & HGD                      & \textbf{3.2} & 5.4           & 6.6           & 35.5          & 12.5          & 12.6          \\
                                  & NRP                      & 5.5          & 6.8           & 9.1           & 12.7          & 13.2          & 9.5           \\ \cmidrule(r){2-8} 
                                  & RDP                     & 3.8          & \textbf{5.2}  & \textbf{5.9}  & \textbf{9.8}  & \textbf{9.8}  & \textbf{6.9}  \\\midrule
\multirow{10}{*}{\rotatebox{90}{Inc-v4}}          & JEPG                     & 23.7         & 27.9          & 41.1          & 41.2          & 52.2          & 37.2          \\
                                  & FD                       & 28.9         & 30.1          & 36.3          & 47.7          & 48.7          & 38.3          \\
                                  & Bit-Red                      & 10.1         & 12.8          & 17.1          & 37.6          & 27.0          & 20.9          \\
                                  & TVM                      & 11.8         & 17.7          & 22.4          & 39.4          & 35.7          & 25.4          \\
                                  & SR                       & 33.7         & 40.2          & 54.5          & 47.1          & 52.1          & 45.5          \\
                                  & R\&P                     & 10.5         & 15.4          & 19.5          & 44.4          & 30.2          & 24.0          \\
                                  & ComDefend                & 24.6         & 25.3          & 30.3          & 45.7          & 44.8          & 34.1          \\
                                  & HGD                      & \textbf{3.5} & 7.3           & 10.2          & 38.7          & 18.5          & 15.6          \\
                                  & NRP                      & 6.9          & 7.6           & 9.3           & 13.0          & 18.0          & 11.0          \\ \cmidrule(r){2-8} 
                                  & RDP                     & 4.3          & \textbf{6.6}  & \textbf{6.5}  & \textbf{9.4}  & \textbf{13.2} & \textbf{8.0}  \\\midrule
\multirow{10}{*}{\rotatebox{90}{ResNet-101}} & JEPG                     & 25.3         & 30.2          & 50.9          & 51.0          & 50.2          & 41.5          \\
                                  & FD                       & 32.7         & 33.3          & 47.3          & 58.0          & 47.6          & 43.8          \\
                                  & Bit-Red                      & 10.3         & 15.2          & 24.0          & 46.7          & 26.1          & 24.5          \\
                                  & TVM                      & 14.1         & 21.4          & 34.8          & 48.2          & 35.1          & 30.7          \\
                                  & SR                       & 35.8         & 38.0          & 57.8          & 46.4          & 49.0          & 45.4          \\
                                  & R\&P                     & 12.7         & 18.8          & 30.3          & 50.2          & 30.9          & 28.6          \\
                                  & ComDefend                & 25.7         & 25.8          & 36.2          & 55.3          & 41.8          & 37.0          \\
                                  & HGD                      & \textbf{4.8} & 12.5          & 18.2          & 47.5          & 20.1          & 20.6          \\
                                  & NRP                      & 9.3          & 11.6          & 15.7          & 22.2          & 18.9          & 15.5          \\ \cmidrule(r){2-8} 
                                  & RDP                     & 6.0          & \textbf{8.9}  & \textbf{10.7} & \textbf{20.6} & \textbf{15.3} & \textbf{12.3} \\\midrule
\multirow{10}{*}{\rotatebox{90}{ResNet-152}}      & JEPG                     & 25.5         & 28.2          & 48.9          & 52.1          & 48.1          & 40.6          \\
                                  & FD                       & 31.1         & 30.1          & 45.1          & 57.5          & 46.6          & 42.1          \\
                                  & Bit-Red                      & 10.5         & 14.9          & 25.3          & 48.8          & 25.9          & 25.1          \\
                                  & TVM                      & 13.1         & 19.2          & 34.2          & 48.3          & 33.0          & 29.6          \\
                                  & SR                       & 34.1         & 35.7          & 57.4          & 45.0          & 47.5          & 43.9          \\
                                  & R\&P                     & 10.9         & 18.0          & 30.2          & 51.2          & 28.1          & 27.7          \\
                                  & ComDefend                & 25.1         & 24.0          & 36.7          & 56.2          & 38.7          & 36.1          \\
                                  & HGD                      & \textbf{4.7} & 10.3          & 18.9          & 48.7          & 18.7          & 20.3          \\
                                  & NRP                      & 8.3          & 10.0          & 16.1          & 21.5          & 17.0          & 14.6          \\ \cmidrule(r){2-8} 
                                  & RDP                     & 5.9          & \textbf{7.7}  & \textbf{11.7} & \textbf{19.9} & \textbf{13.3} & \textbf{11.7} \\ \midrule
\multirow{10}{*}{\rotatebox{90}{ensemble}}        & JEPG                     & 31.0         & 57.3          & 74.1          & 69.5          & 61.9          & 58.8          \\
                                  & FD                       & 34.9         & 50.3          & 64.6          & 68.6          & 52.6          & 54.2          \\
                                  & Bit-Red                      & 12.6         & 26.4          & 38.5          & 65.0          & 26.9          & 33.9          \\
                                  & TVM                      & 16.2         & 35.8          & 51.2          & 66.6          & 35.3          & 41.0          \\
                                  & SR                       & 38.8         & 61.0          & 82.1          & 65.2          & 61.3          & 61.7          \\
                                  & R\&P                     & 13.1         & 32.1          & 48.4          & 70.2          & 29.3          & 38.6          \\
                                  & ComDefend                & 30.5         & 42.7          & 57.1          & 68.8          & 45.1          & 48.8          \\
                                  & HGD                      & \textbf{4.0} & 17.1          & 25.5          & 68.2          & 16.0          & 26.2          \\
                                  & NRP                      & 11.1         & 18.5          & 25.9          & 31.5          & 18.1          & 21.0          \\ \cmidrule(r){2-8} 
                                  & RDP                     & 5.5          & \textbf{13.1} & \textbf{16.7} & \textbf{26.2} & \textbf{13.0} & \textbf{14.9}    \\ \bottomrule
\end{tabular}
}
\end{table}

\begin{table}[t]
\centering
\caption{Clean accuracy (higher is better) of different defenses. We use \textcolor{red}{$\uparrow$} to indicate the results which are higher than the original accuracy of IncRes-v2\textsubscript{ens} on clean images.}
\label{tab:clean acc}
\resizebox{\columnwidth}{!}{%
\begin{tabular}{cccccc}
\toprule
\textbf{Defense} & JEPG      & FD      & BIT              & TVM     & SR             \\ \midrule
\textbf{Acc}     & 97.40\%   & 90.30\% & 97.20\%          & 96.20\% & 97.70\% \textcolor{red}{$\uparrow$}          \\ \toprule
\textbf{Defense} & R\&P & ComDefend & HGD     & NRP     & RDP             \\ \midrule
\textbf{Acc}     & 97.10\% & 90.50\%   & 97.20\% & 94.90\% & 97.70\% \textcolor{red}{$\uparrow$} \\\bottomrule
\end{tabular}
}
\end{table}

\begin{table}[h!]
\centering
\caption{Success rates (higher is better) of RDP against BPDA and TI-DIM attacks.}
\label{tab:bpda}
\resizebox{\columnwidth}{!}{%
\begin{tabular}{cccc}
\toprule
\multirow{4}{*}{\diagbox[]{Target}{Source}}
                     & ResNet-152  & ResNet-152  & \makecell[c]{ResNet-152\\ NRP}  \\ \cmidrule(r){2-4}
               & TI-DIM  & TI-DIM  & \makecell[c]{TI-DIM \\ BPDA} \\ \midrule
With RDP              & \xmark & \cmark     & \cmark           \\ \midrule
Inc-v3               & 37.10\% & 71.50\% & 64.60\%       \\ 
Inc-v4               & 43.10\% & 75.40\% & 71.90\%       \\
Incres-v2            & 45.60\% & 83.60\% & 83.10\%       \\
IncRes-v2\textsubscript{ens}        & 50.20\% & 80.80\% & 79.20\%      \\ \bottomrule
\end{tabular}
}
\end{table}

\begin{table}[h!]
\centering
\caption{Defense success rates~(higher is better) against PGD-10 attack in white-box settings on ImageNet validation set (50k images). ResNet-152 is used as the backbone.}
\label{tab:acc imagenet 50k}
\resizebox{0.95\columnwidth}{!}{
\begin{tabular}{ccccc}
\toprule
\textbf{Defense}      & N/A & AT-FD & NRP     & RDP     \\
\midrule
Clean       & \textbf{82.41\%}  & 65.30\%           & 73.86\% & 81.40\% \\
Adversarial & 0.74\%   & 55.70\%           & 65.60\% & \textbf{71.30\%} \\
\bottomrule
\end{tabular}
}
\end{table}

\begin{figure}[h!]
\centering
\includegraphics[width=\linewidth]{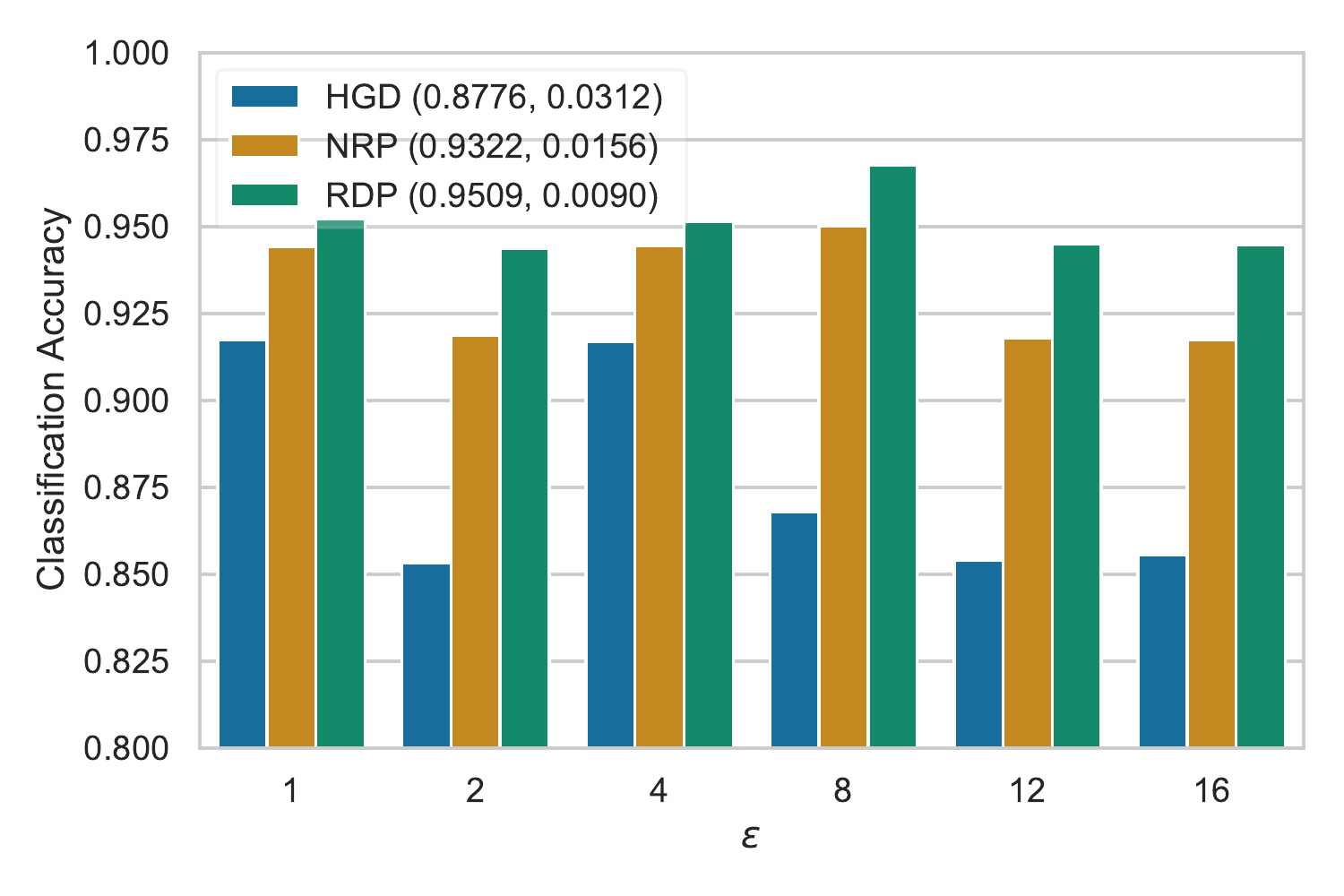}
\caption{
Success rates of HGD, NRP and RDP when defending against various attacks with $\epsilon=1, 2, 4, 8, 12, 16$. 
The mean and standard deviation of each method are behind their names.
We use Inc-v3 as the source model for attacks and IncRes-v2 as the backbone.
}
\label{fig:multi-eps}
\end{figure}

\subsection{Evaluation Results}

\begin{figure*}[ht!]
\centering
\begin{subfigure}{0.19\linewidth}
  \includegraphics[width=\linewidth]{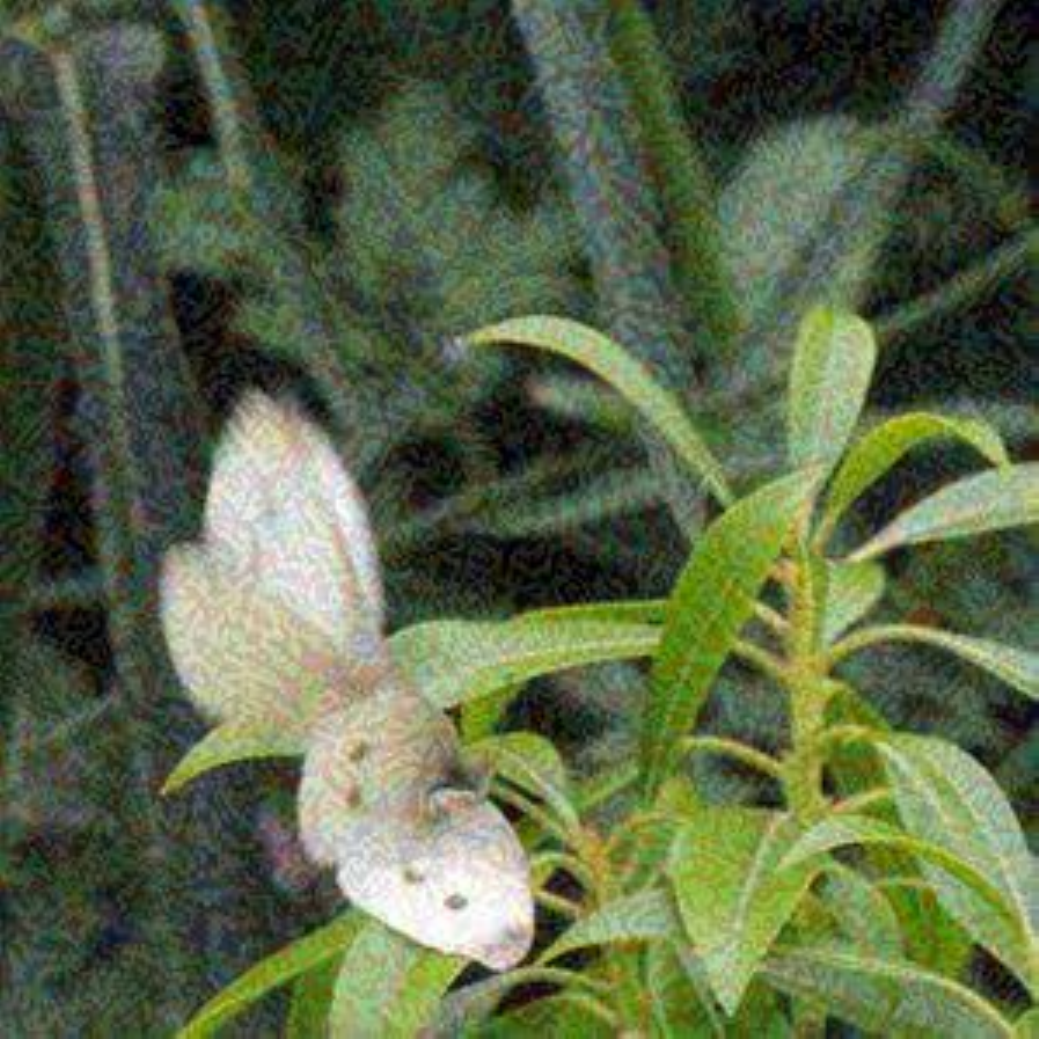}
\caption*{{\fontsize{8pt}{12pt}\selectfont FGSM:} bullfrog (\xmark)}
\end{subfigure}
\begin{subfigure}{0.19\linewidth}
\includegraphics[width=\linewidth]{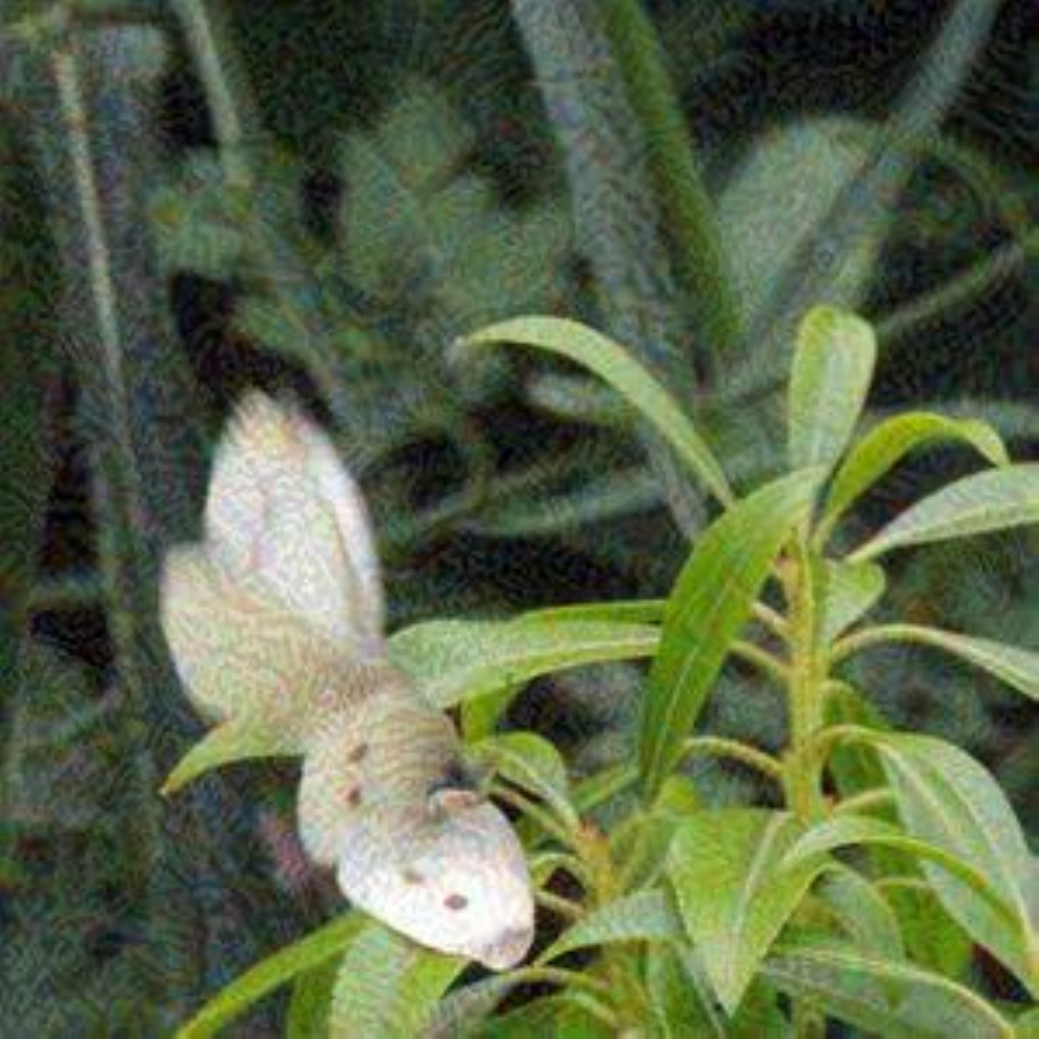}
\caption*{{\fontsize{8pt}{12pt}\selectfont MI-FGSM:} bullfrog (\xmark)}
\end{subfigure}
\begin{subfigure}{0.19\linewidth}
\includegraphics[width=\linewidth]{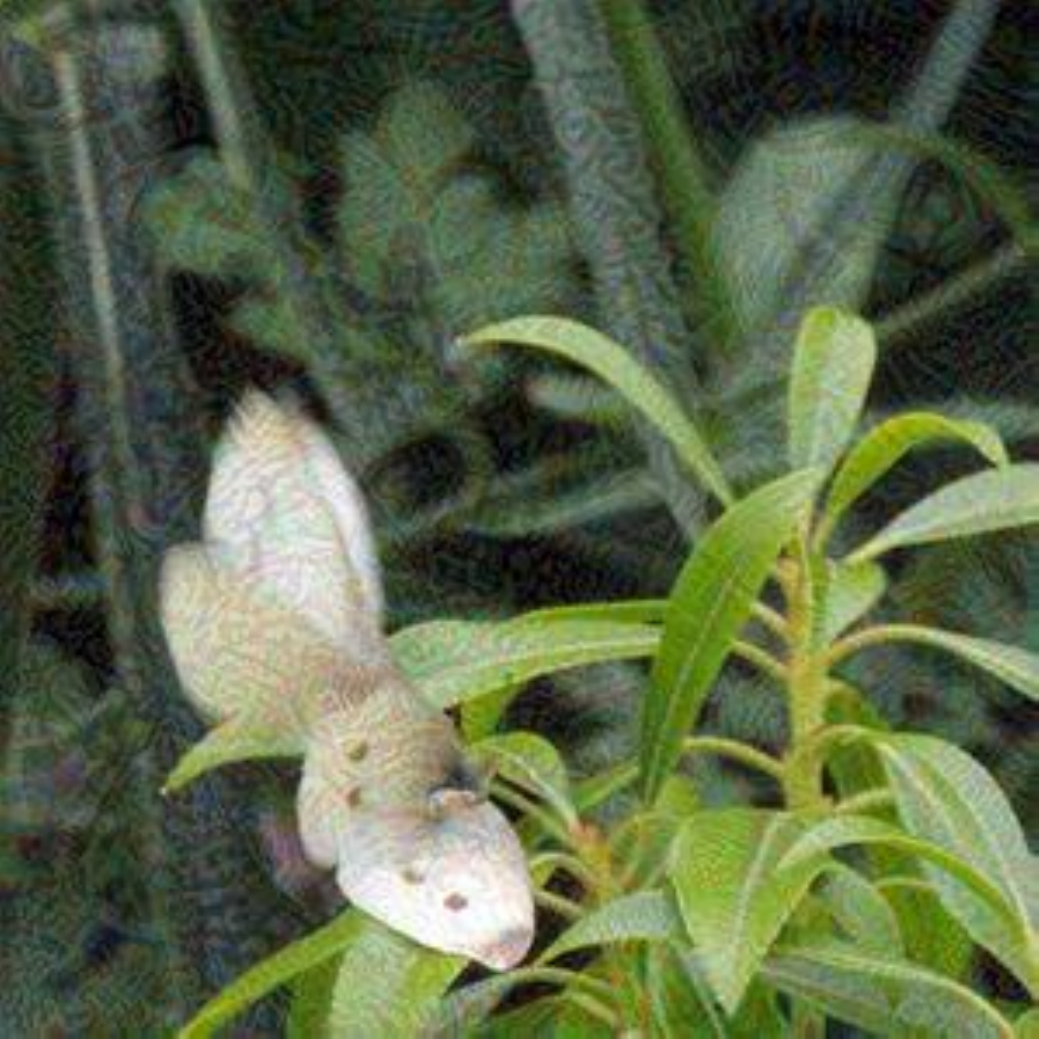}
\caption*{{\fontsize{8pt}{12pt}\selectfont DIM:} bullfrog (\xmark)}
\end{subfigure}
\begin{subfigure}{0.19\linewidth}
\includegraphics[width=\linewidth]{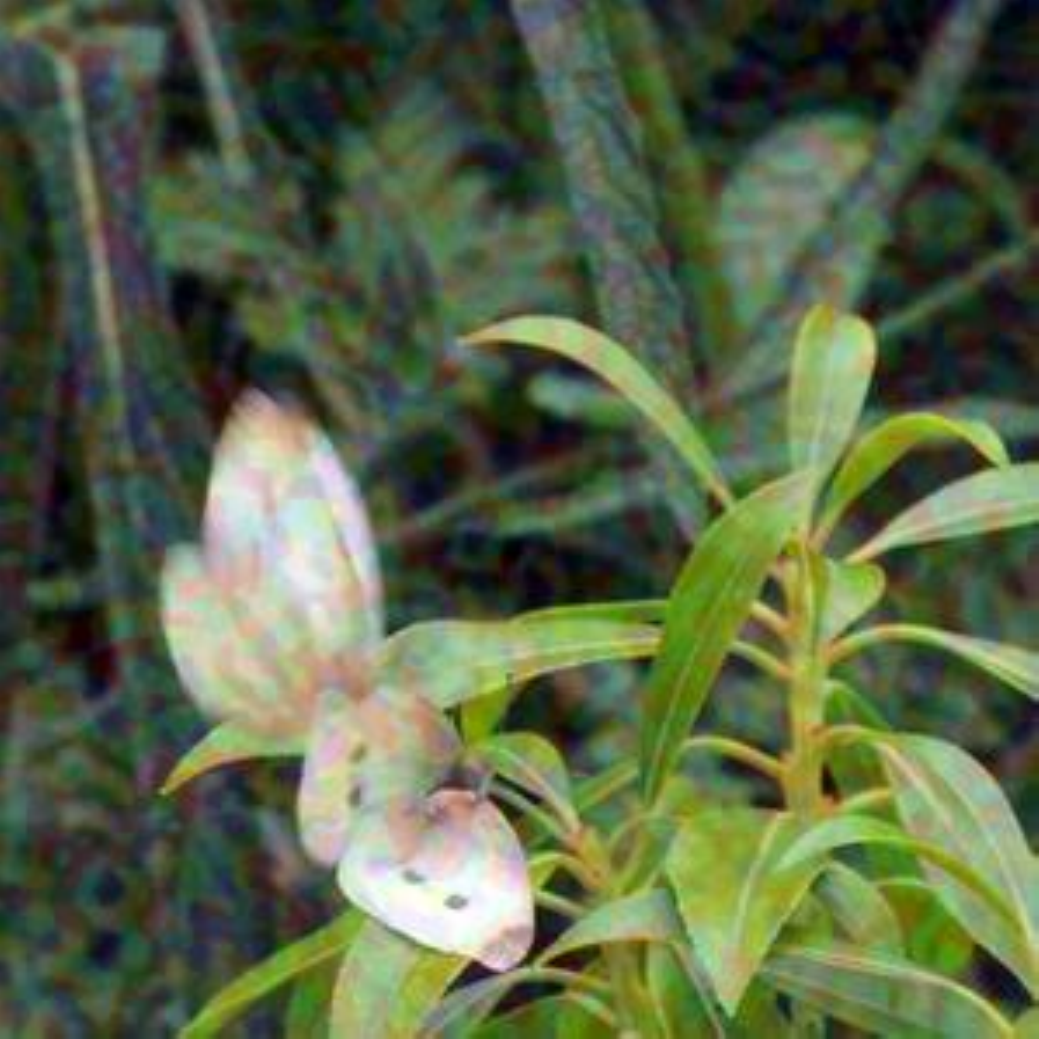}
\caption*{{\fontsize{8pt}{12pt}\selectfont TI-DIM:} corn (\xmark)}
\end{subfigure}
\begin{subfigure}{0.19\linewidth}
\includegraphics[width=\linewidth]{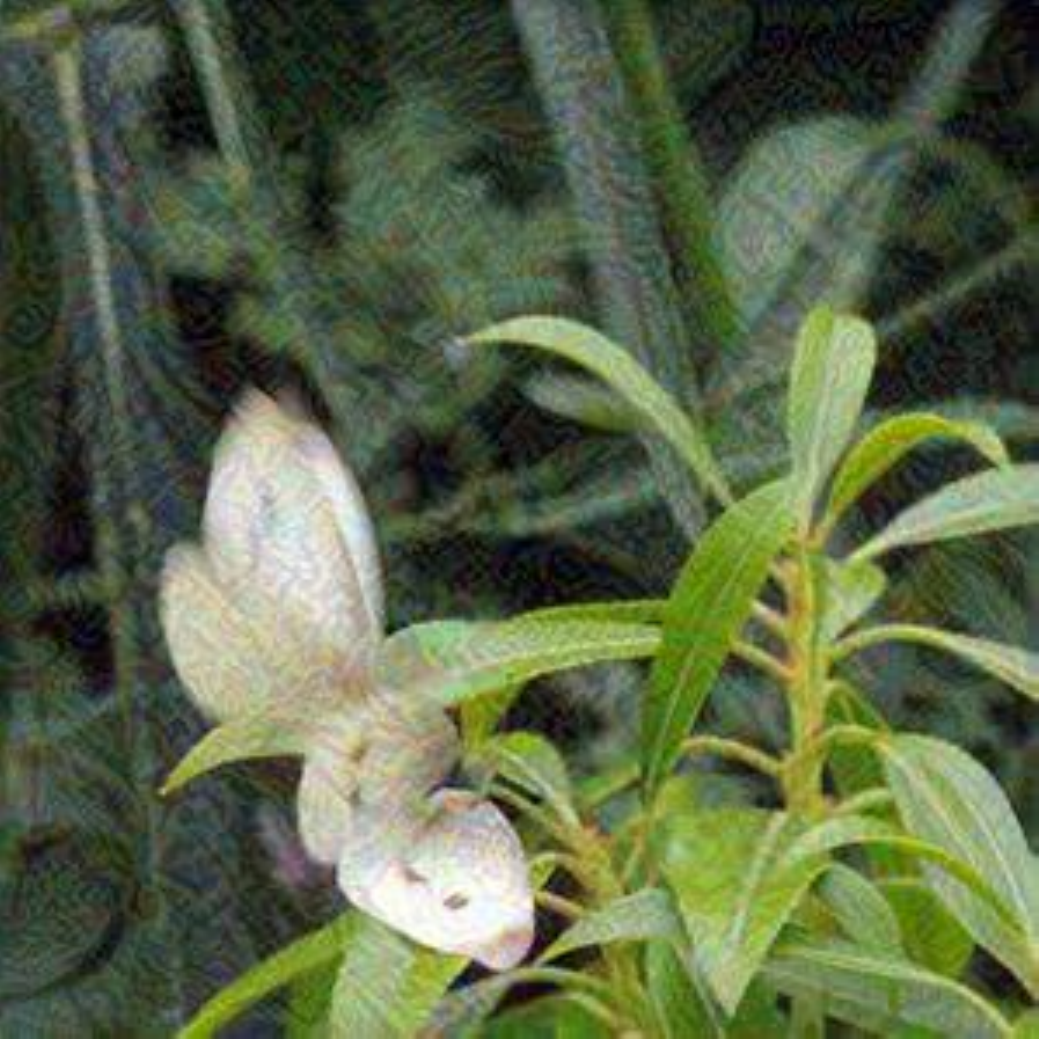}
\caption*{{\fontsize{8pt}{12pt}\selectfont SINI-TIDIM:} kingsnake (\xmark)}
\end{subfigure}
\vspace{.4em}
\begin{subfigure}{0.19\linewidth}
\caption*{Sulphur butterfly (\cmark)}
\includegraphics[width=\linewidth]{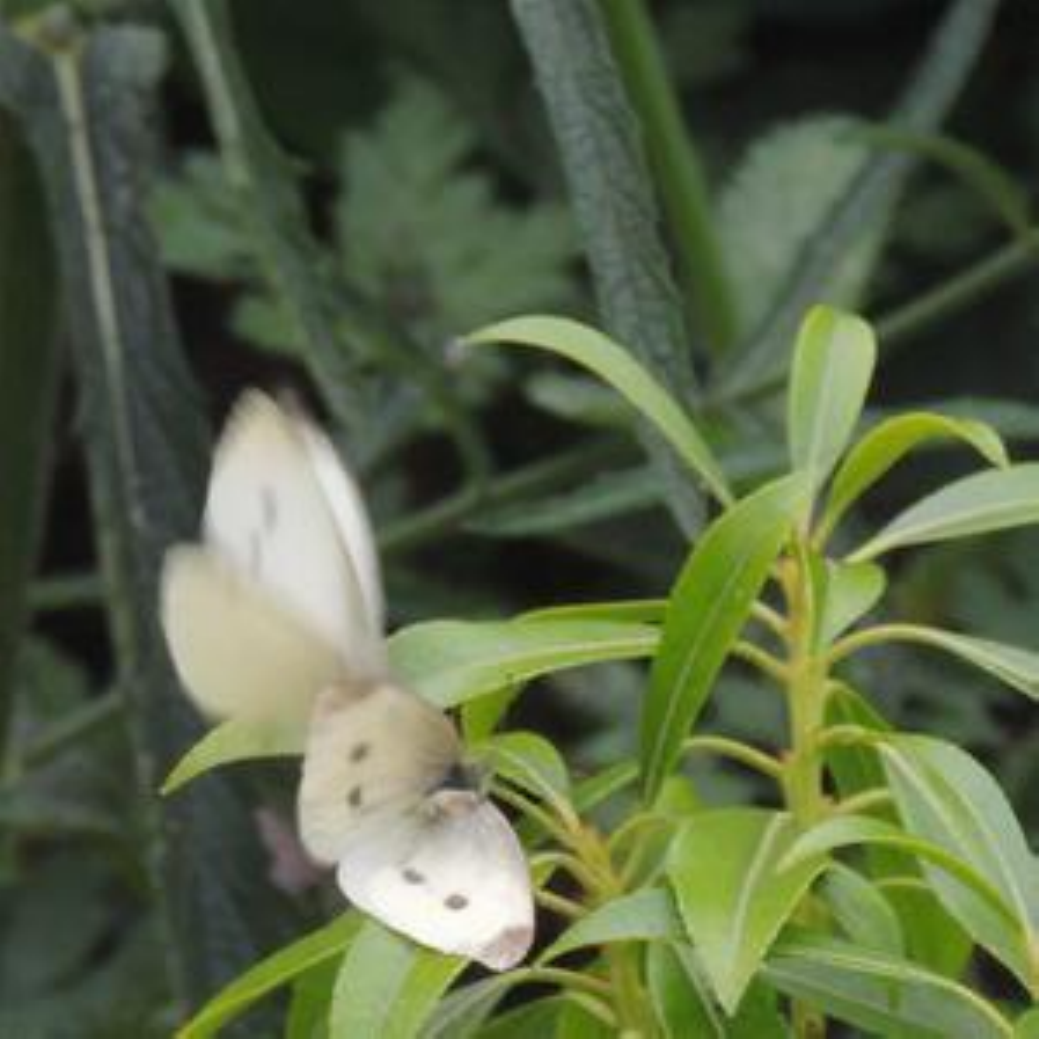}
\end{subfigure}
\begin{subfigure}{0.19\linewidth}
\caption*{Sulphur butterfly (\cmark)}
\includegraphics[width=\linewidth]{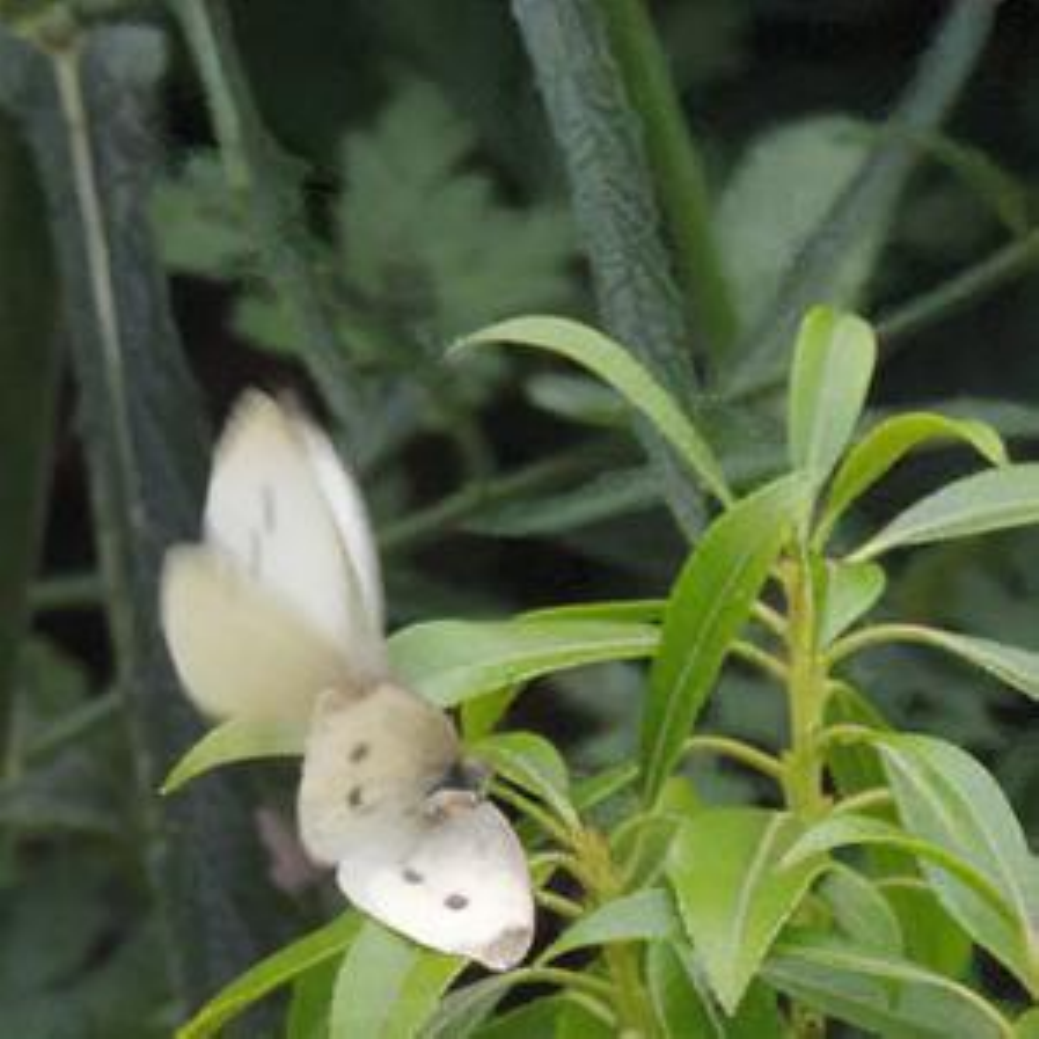}
\end{subfigure}
\begin{subfigure}{0.19\linewidth}
\caption*{Sulphur butterfly (\cmark)}
\includegraphics[width=\linewidth]{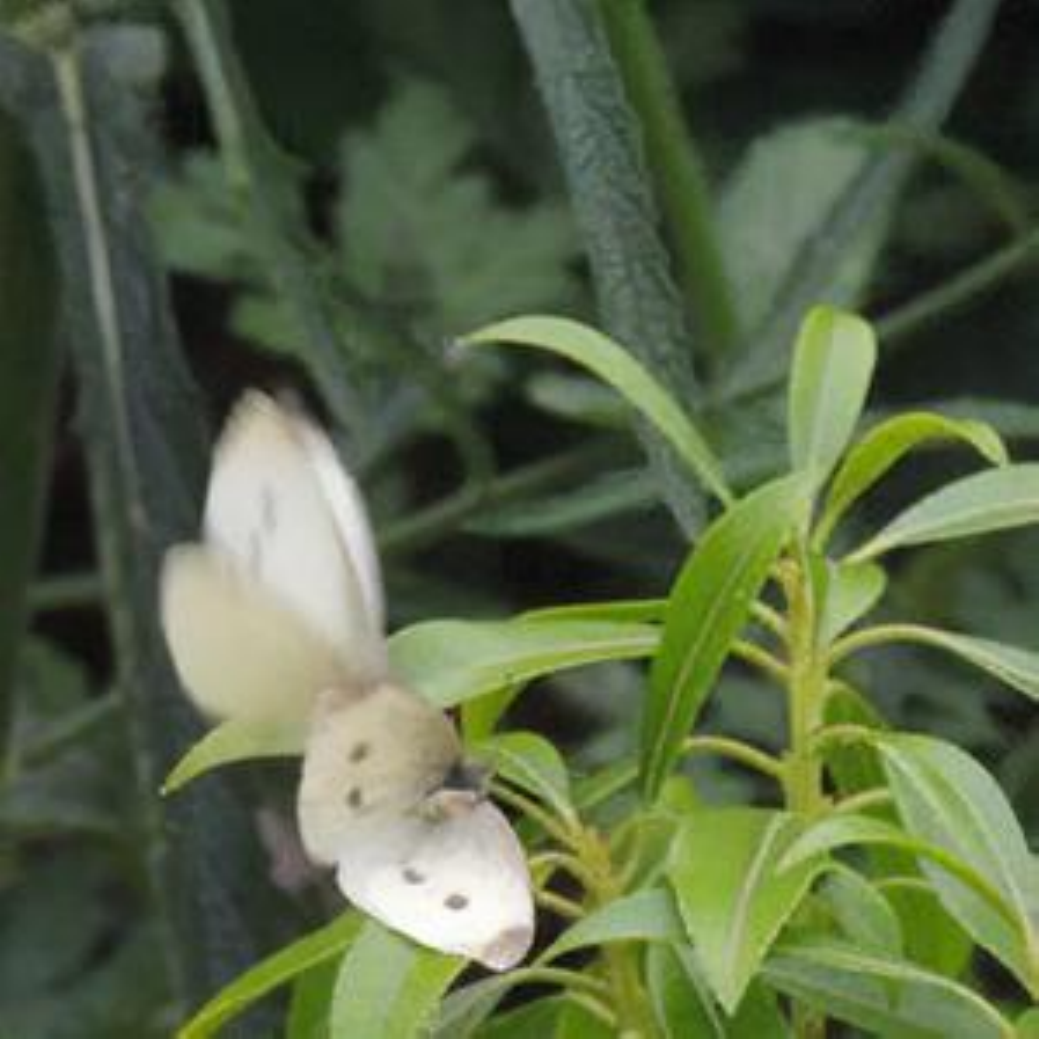}
\end{subfigure}
\begin{subfigure}{0.19\linewidth}
\caption*{Sulphur butterfly (\cmark)}
\includegraphics[width=\linewidth]{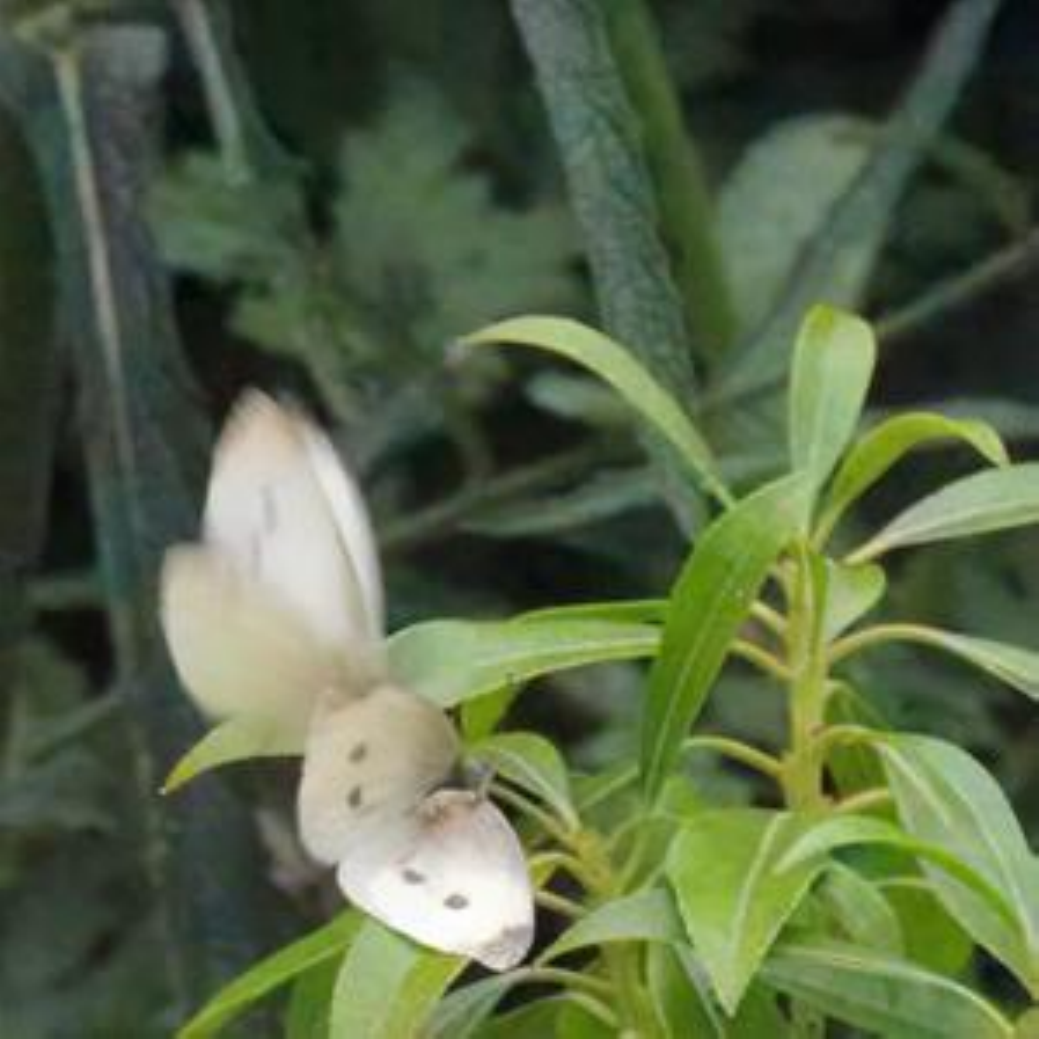}
\end{subfigure}
\begin{subfigure}{0.19\linewidth}
\caption*{Sulphur butterfly (\cmark)}
\includegraphics[width=\linewidth]{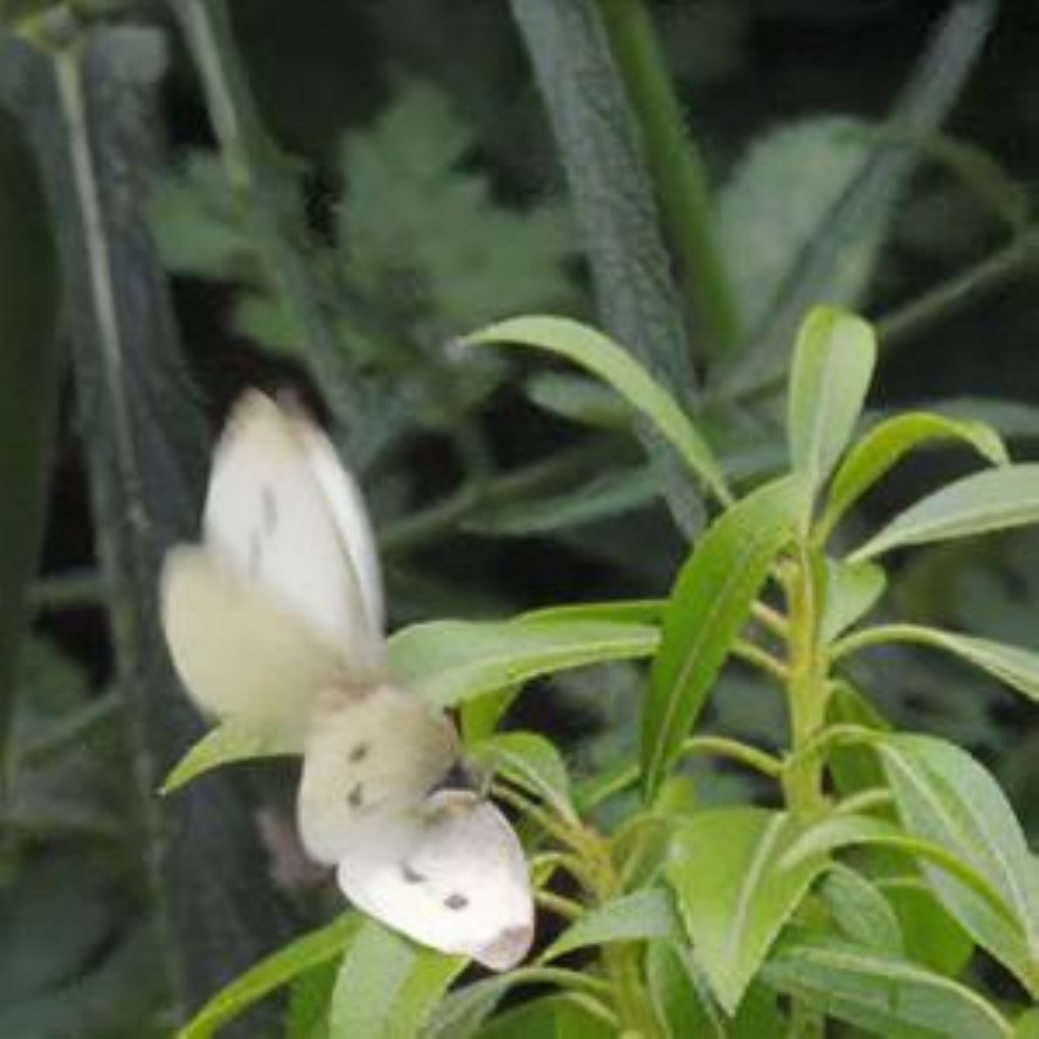}
\end{subfigure}
\caption{
An illustration of the generalizability of RDP against various strong attacks ($\epsilon=16$). Images on the top row are generated by different adversarial attacks, while images on the bottom row are purified by RDP. RDP effectively removes the adversarial patterns and maintains the samples in high quality. we use \cmark and \xmark~to indicate the correctness of predictions, where IncRes-v2\textsubscript{ens} is used as the backbone.
}
\label{fig:defenses vis}
\end{figure*}

\subsubsection{Defending various attacks}
Compared to adversarial training, one advantage of preprocessing-based defenses is that there is no assumption of the potential attacks and downstream tasks.
To demonstrate the generalization ability, RDP is deployed to defend against various strong attacks like FGSM, MI-FGSM, DIM, TI-DIM and SINI-TIDIM, where $\epsilon \in [0,255]$ and is set as 16 under $l_{\infty}$ norm, four naturally trained models and their ensembles are used as the source models.
We also compare RDP with previous representative methods like JPEG compression~\cite{DBLP:journals/corr/abs-1711-00117}, TVM~\cite{DBLP:journals/corr/abs-1711-00117}, Bit Depth Reduction~(Bit-Red)~\cite{xu2017feature}, Feature Distillation~(FD)~\cite{liu2019feature}, SR~\cite{mustafa2019image}, R\&P~\cite{xie2018mitigating}, ComDefend~\cite{jia2019comdefend} and the state-of-the-art methods like HGD~\cite{DBLP:conf/cvpr/LiaoLDPH018} and NRP~\cite{Naseer_2020_CVPR}.
All these methods are originally developed to defend adversarial examples with $\epsilon=16$ on ImageNet.
IncRes-v2\textsubscript{ens} is used as the backbone model here.
Quantitative results are summarized in Table~\ref{tab:various attacks}.
As we can see, among those methods, RDP holds the strongest robustness when defending most attacks.
HGD performs better than RDP when defending FGSM, the reason of which is that the training data of HGD contains a large portion of adversarial examples generated by FGSM~\cite{DBLP:conf/cvpr/LiaoLDPH018} while we don't assume the potential attacks when training RDP.
Compared to the SOTA method NRP, RDP holds an improvement of $3.6\%$ on average.
In addition, RDP doesn't hurt the performances of the backbone model on clean data while other defenses sacrifice the accuracy~(97.60\%) of the backbone model up to 7.10\% (see Table~\ref{tab:clean acc}).

\subsubsection{Defending attacks with varying $\epsilon$}
The magnitude $\epsilon$ of attacks is usually fixed for both training and testing in the literature~\cite{DBLP:conf/cvpr/LiaoLDPH018,Naseer_2020_CVPR}.
Nevertheless, we argue that this could conceal the overfitting issue of learning-based defenses.
The potential risk of being overfitting would make defenses unreliable to weak attacks.
To address this concern, we prepare another evaluation set consisting of adversarial examples generated by the same attacks but with varying $\epsilon=\{1,2,4,8,12,16\}$, which is missing in the literature.
Inc-v3 is used here as the target model for attacks.
We mainly choose HGD and NRP for comparison, which are the most relevant to RDP and have good performances in previous experiments.
As illustrated in Figure~\ref{fig:multi-eps}, across various $\epsilon$, the mean value of classification accuracies defended by RDP is highest, and the standard deviation is the smallest.
To conclude, RDP generalizes well across various attacks with different $\epsilon$, the performances of which are more stable than NRP and HGD.

\subsubsection{Adversaries' Awareness of Defenses}
In the above experiments, we assume the adversaries have no idea about the defenses.
However, as suggested in~\cite{athalye2018obfuscated}, adversaries could approximate the gradients to evade defenses when defenses are deployed.
Here, we investigate the performances of RDP in this difficult scenario.
We make two primary assumptions of the adversaries: (\romannum{1}) the adversaries are aware of the existence of our defense and have access to the exact or similar training data, (\romannum{2}) the adversaries train a similar defense and combines it with BPDA~\cite{athalye2018obfuscated} and TI-DIM attack to bypass our defense.
Here we use the current SOTA method: NRP~\cite{Naseer_2020_CVPR} as the adversaries' local defense and ResNet-152 as the source model to simulate this attack.
Three naturally-trained models and one adversarially-trained model are used for evaluation.
The experimental results are summarized in Table~\ref{tab:bpda}.
For attacks with BPDA, the performances of RDP only drop slightly.
What's more, with RDP as the defense, the average relative gain of four target models is 74\%.
To sum up, RDP can effectively defend against adversarial attacks and is resistant to BPDA.

Further, we consider the worst scenario where adversaries can access both RDP and the backbone model when launching attacks.
Following~\cite{DBLP:conf/cvpr/XieWMYH19}, we test RDP under the targeted PGD-10 attack with $\epsilon=16$ in white-box settings.
For hardening our defense, we propose to make RDP dynamic by incorporating random smoothing strategy~\cite{DBLP:conf/icml/CohenRK19,Naseer_2020_CVPR}.
Specifically, we add one Gaussian noise layer (standard deviation is 0.05) on the top of RDP when inference.
In Table~\ref{tab:acc imagenet 50k}, we demonstrate the efficacy of RDP defending against white-box attacks.
RDP significantly improves the robustness of the backbone model while maintaining the clean accuracy, which is apparently superior to adversarial training with feature denoising (AT-FD)~\cite{DBLP:conf/cvpr/XieWMYH19}.

\begin{figure}[t]
\begin{subfigure}[l]{0.91\linewidth}
\includegraphics[width=\linewidth]{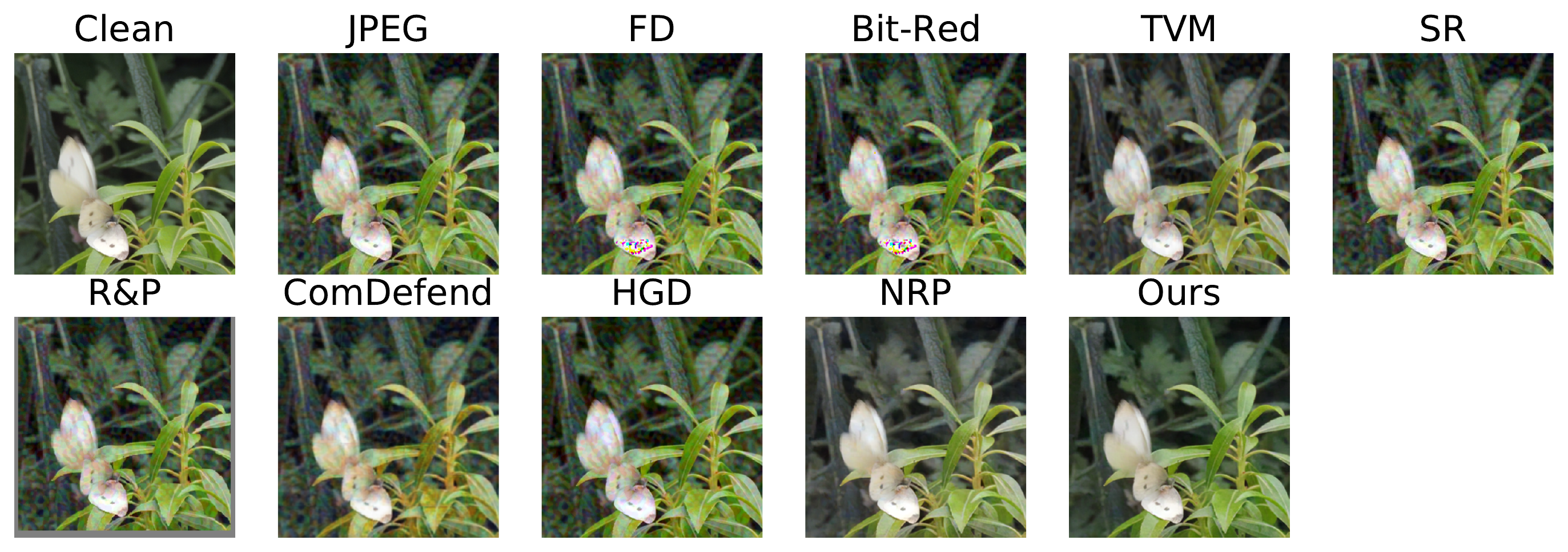}
\end{subfigure}
\vspace{.4em}
\begin{subfigure}[l]{\linewidth}
\includegraphics[width=\linewidth]{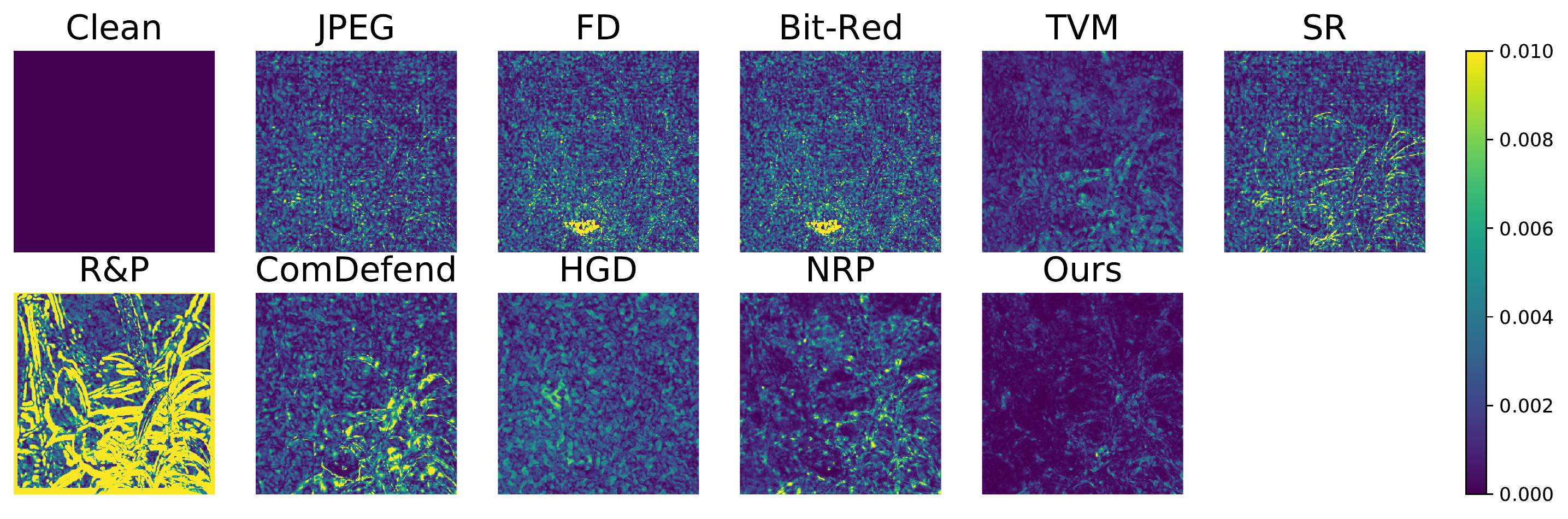}
\end{subfigure}
\caption{
Restoration results (upper) and the magnitudes of the corresponding errors (lower) by different defense methods. The darker, the better. 
Adversarial example purified by RDP preserves the highest similarity with the clean image.
}
\label{fig:error}
\end{figure}

\subsection{Disentangled Outputs by RDP}
As illustrated in Figure~\ref{fig:arch}, RDP has two branches: purification and reconstruction. 
When defending against adversarial attacks, we only use the purification branch.
Samples of different adversarial examples and their corresponding purified images by RDP are shown in Figure~\ref{fig:defenses vis}.
We also show the processed adversarial examples by other defenses in Figure~\ref{fig:error}.
By visualizing the error maps between the processed images and clean images, we could see that RDP has the best restoration ability, which is also reflected in the strong adversarial robustness of RDP.

\begin{figure}[t]
\centering
\begin{subfigure}{0.24\columnwidth}
\includegraphics[width=\columnwidth]{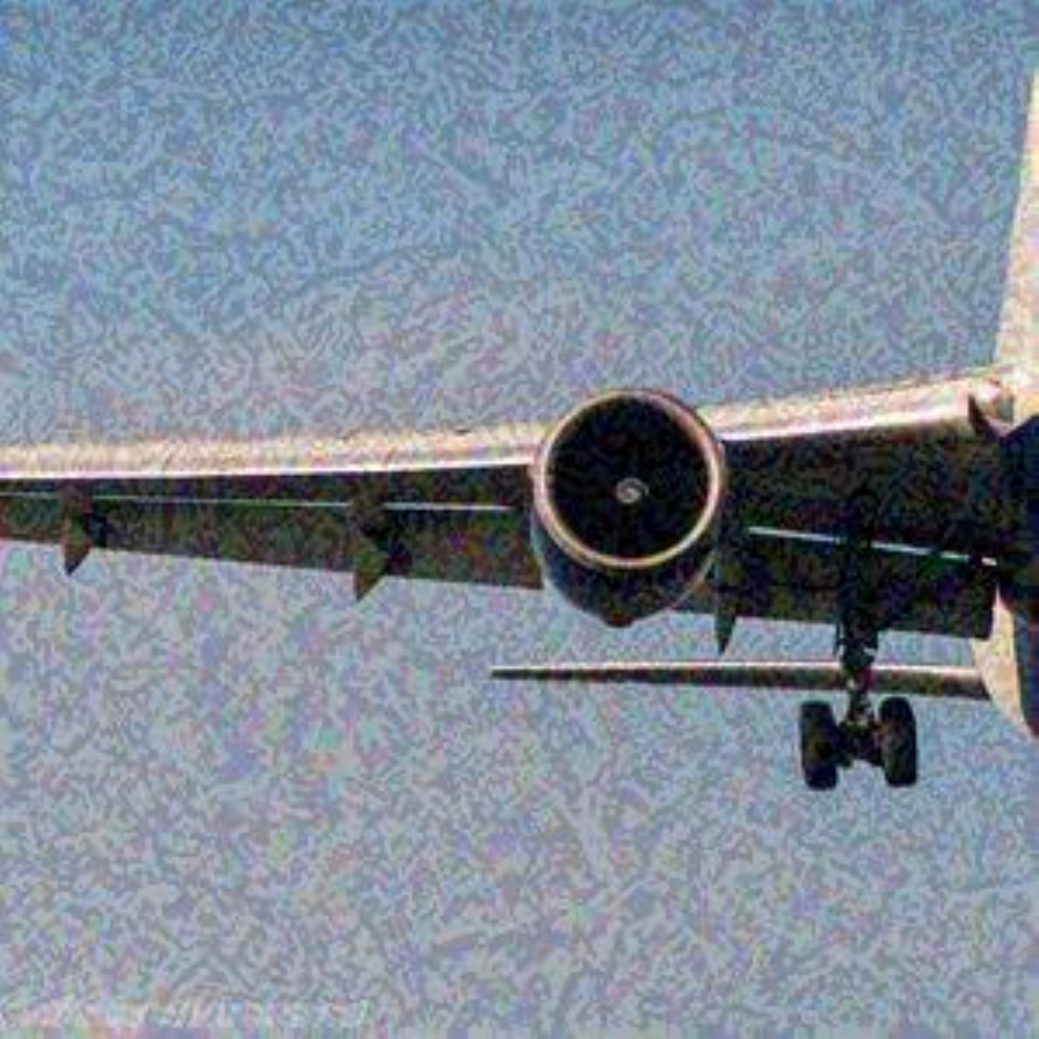}
\caption{}\label{fig:disentangled a}
\end{subfigure}
\begin{subfigure}{0.24\columnwidth}
\includegraphics[width=\columnwidth]{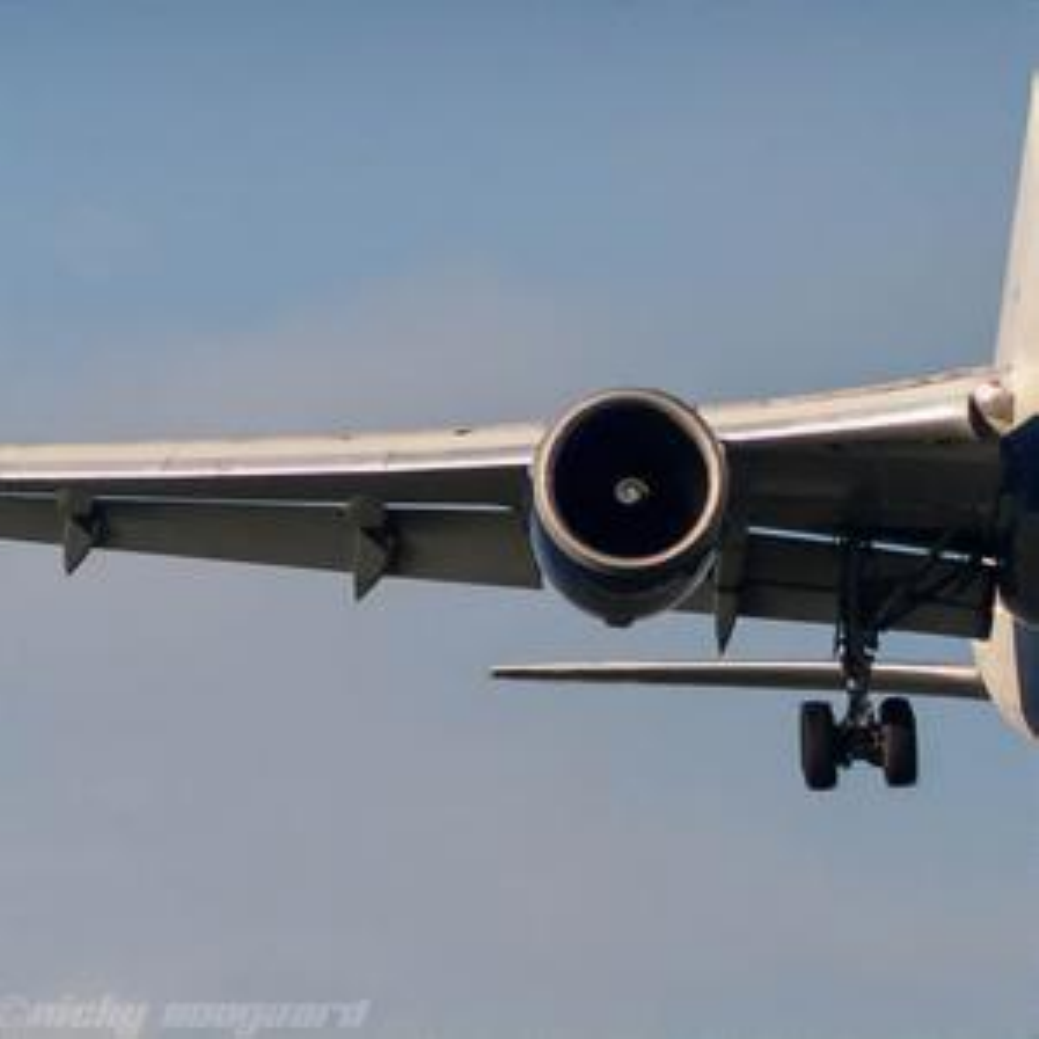}
\caption{}\label{fig:disentangled b}
\end{subfigure}
\begin{subfigure}{0.24\columnwidth}
\includegraphics[width=\columnwidth]{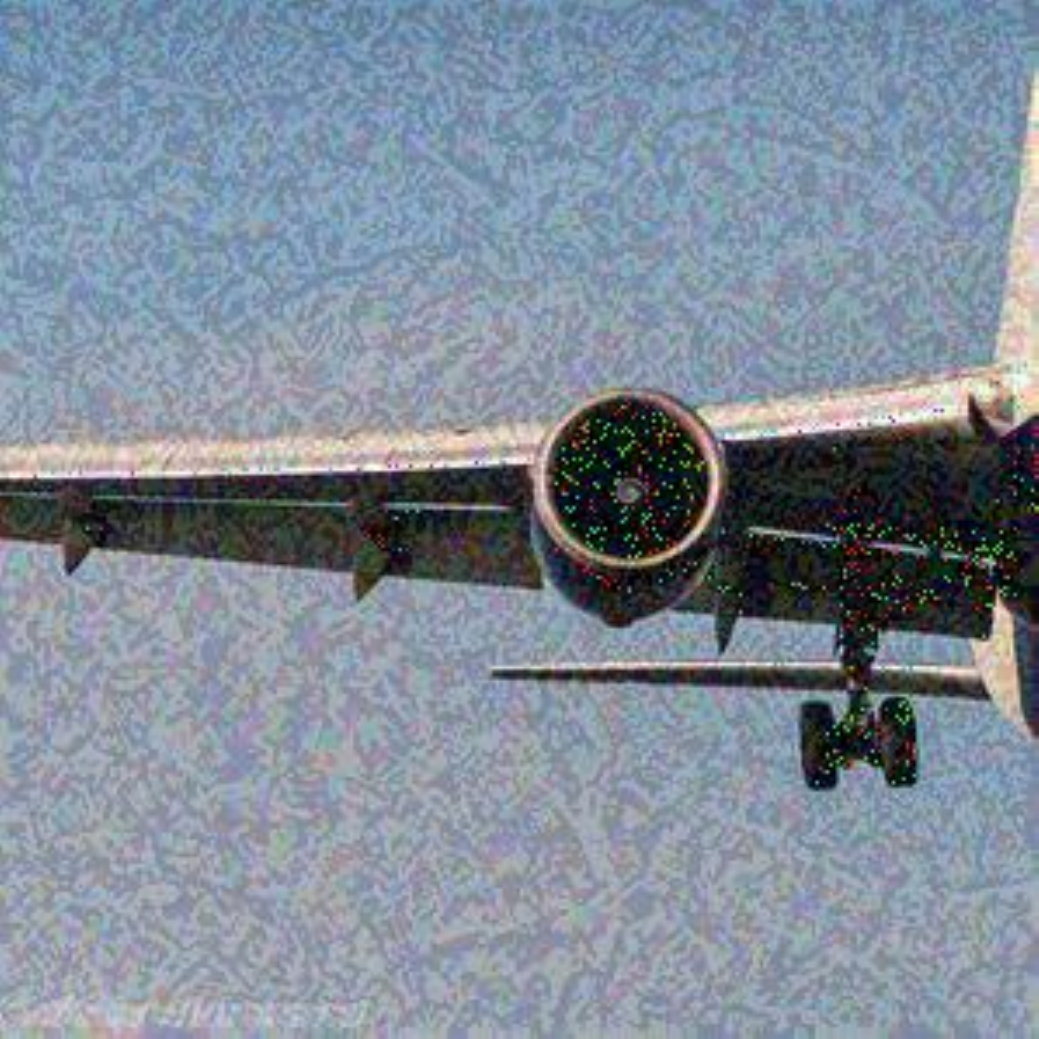}
\caption{}\label{fig:disentangled c}
\end{subfigure}
\begin{subfigure}{0.24\columnwidth}
\includegraphics[width=\columnwidth]{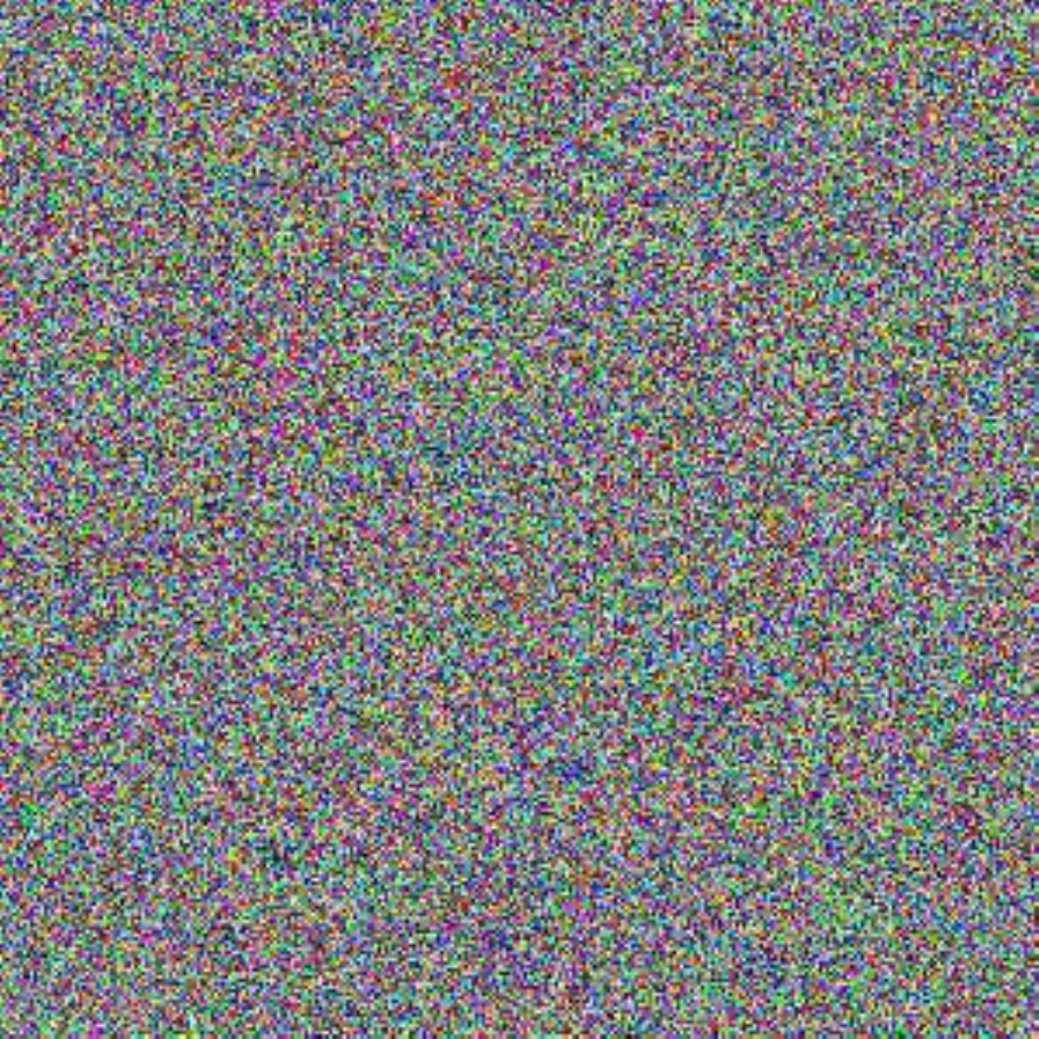}
\caption{}\label{fig:disentangled d}
\end{subfigure}
\caption{
Disentangled outputs of RDP. (a) is the adversarial example. (b) and (c) are the outputs of purification and reconstruction branches, respectively. (d) is the output of reconstruction branch when replacing the natural latent vector with random noise.
}
\label{fig:disentangled}
\end{figure}

For better understanding the usage of the reconstruction branch in RDP, given an adversarial example~(Figure~\ref{fig:disentangled a}), we visualize and show the disentangled outputs of two branches in Figure~\ref{fig:disentangled b} and~\ref{fig:disentangled c}, respectively.
Further, we fix the reconstruction branch's latent vector and replace the purification branch's latent vector with random noise.
The new reconstruction output is shown in Figure~\ref{fig:disentangled d}, which indicates that RDP does disentangle the contents and the perturbations of adversarial examples.

\subsection{Ablation Study}
\begin{figure}[t]
\centering
\includegraphics[width=\linewidth]{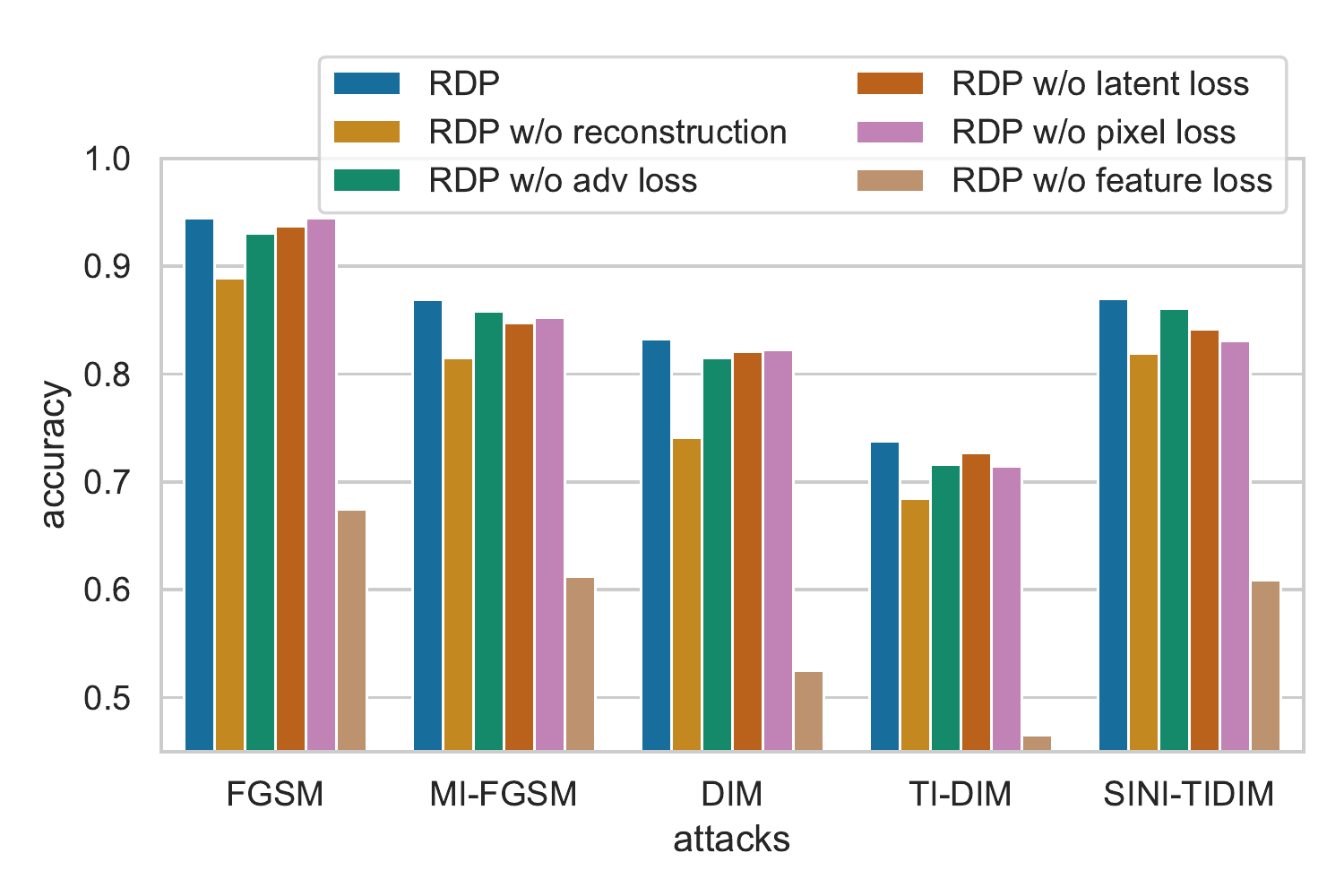}
\caption{
Ablation study of RDP. Results are obtained by IncRes-v2\textsubscript{ens} against the \textbf{ensemble}
attacks in previous experiments.}
\label{fig:ablation}
\end{figure}

Here, we thoroughly investigate the impact of disentanglement training and the use of different objective functions with our defense.
Figure~\ref{fig:ablation} shows the evaluation results and provides the following conclusions: 
(\romannum{1}) the reconstruction branch indeed helps the purification of adversarial examples;
(\romannum{2}) Performances of RDP without latent loss, pixel loss, and adversarial loss drop slightly;
(\romannum{3}) Without feature loss, RDP cannot project adversarial examples to the manifold of natural images, and the defense gets much weaker.

\section{Conclusion}
This paper proposes a novel defense method called Representation Disentangled Purification (RDP) to defend against adversarial attacks.
By disentangling the image content and adversarial patterns, RDP effectively recovers the clean images in high quality.
With extensive experiments, RDP exhibits high generalizability to the unseen strong attacks and outperforms the existing SOTA defense methods by a large gap.
The high generalizability makes RDP potentially be applied to many kinds of downstream attacks and provides high adversarial robustness.
Incorporating with the dynamic strategy, RDP is also resistant to strong white-box attacks without sacrificing the clean accuracy of backbone models.

\bibliography{ref.bib}

\end{document}